\documentclass{article}
\usepackage{ijcai22}

\usepackage{times}
\usepackage{soul}
\usepackage{url}
\usepackage[hidelinks]{hyperref}
\usepackage[utf8]{inputenc}
\usepackage[small]{caption}
\usepackage{graphicx}
\usepackage{amsmath}
\usepackage{amsthm}
\usepackage{booktabs}
\usepackage{algorithm}
\usepackage{algorithmic}
\usepackage{amssymb}
\usepackage{textcomp}
\usepackage{bbold}
\usepackage{multirow}
\usepackage{bm}
\usepackage{adjustbox}
\usepackage{bbold}
\usepackage{xcolor}
\usepackage{subfig}

\title{Empirical Bayesian Approaches for Robust Constraint-based Causal Discovery under Insufficient Data}

\author{
Zijun Cui$^1$
\and
Naiyu Yin$^1$\and
Yuru Wang\footnote{Corresponding author}$^{2}$\And
Qiang Ji$^1$\\
\affiliations
$^{1}$Rensselaer Polytechnic Institute\\
$^2$Northeast Normal University\\
}

\begin{document}

\maketitle

\begin{abstract}
    Causal discovery is to learn cause-effect relationships among variables given observational data and is important for many applications. Existing causal discovery methods assume data sufficiency, which may not be the case in many real world datasets. As a result, many existing causal discovery methods can fail under limited data. In this work, we propose Bayesian-augmented frequentist independence tests to improve the performance of constraint-based causal discovery methods under insufficient data: 1) We firstly introduce a Bayesian method to estimate mutual information (MI), based on which we propose a robust MI based independence test; 2) Secondly, we consider the Bayesian estimation of hypothesis likelihood and incorporate it into a well-defined statistical test, resulting in a robust statistical testing based independence test. We apply proposed independence tests to constraint-based causal discovery methods and evaluate the performance on benchmark datasets with insufficient samples. Experiments show significant performance improvement in terms of both accuracy and efficiency over SOTA methods. 
\end{abstract}

\section{Introduction}

Learning causal relations has been a fundamental and widely-investigated topic. The causal relations are captured by a directed acyclic graph (DAG), and a directed link in DAG captures cause-effect relation between two variables connected by the link~\cite{glymour2019review}. Specifically, a directed link from node $X$ to node $Y$ indicates the cause-effect relation between cause variable $X$ and effect variable $Y$. 
Causal discovery aims at learning a DAG capturing causal-effect relationships among a set of random variables from observational data, and one of the dominant approaches for causal discovery is through the structure causal model (SCM). Existing causal discovery methods focus on learning a DAG with high confidence from sufficient data samples~\cite{yu2019dag}.
Not much attention, however, has been paid to performance improvement of causal discovery under limited data. Such work is important, as even in the era of big data, there are still domains in which the availability of data is very limited. For example, in biological or clinical disciplines, data can be severely insufficient either because of high cost or
lack of cases from which data is collected~\cite{mukherjee2007markov}. 
Furthermore, even for applications with a vast amount of data, the data may not adequately cover all possible states of the nodes, leading to insufficient data for certain states. For example, the observed data under the absence of earthquake is adequate, while the observed data under the occurrence of earthquake is limited, due to the fact that earthquake rarely happens in nature. 

In this paper, we employ constraint-based methods to learn a DAG through independence tests from observational data. Constraint-based causal discovery methods 
can be performed globally or locally.
Global approaches aim at learning cause-effect relationships among all random variables, 
such as PC-stable~\cite{colombo2014order}, and Sepset consistent PC (SC-PC)~\cite{li2019constraint}. 
Global causal discovery methods discussed above learn DAGs that are in the same markov equivalent class of ground truth DAG. 
Further tests under certain assumptions about the graph or data distribution are needed to resolve the causal ambiguity~\cite{glymour2019review}. In this paper, we focus on learning markov equivalent DAGs. In contrast to global approaches, local approaches identify the direct causes and effects of a target variable, represented by a causal Markov Blanket~\cite{gao2015local,yang2021efficient}. A causal Markov Blanket captures local relationships of a target variable by identifying its parents, children, and spouses. For both global and local approaches, the main challenge of constraint-based causal discovery methods is that their performance highly depends on the accuracy of the independence test. 
Independence test error, even one mistake in independence decision, can propagate throughout the graph, causing a sequence of errors and resulting in an erroneous DAG with incorrect orientations~\cite{spirtes2010introduction}. Hence, to perform a robust constraint-based causal discovery, it is crucial to improve the robustness of the independence test. 


To improve the causal discovery performance under insufficient data, we propose to introduce Bayesian approaches to independence tests for accurate and efficient constraint-based causal discovery. Specifically, two Bayesian-augmented frequentist independence tests are proposed, whereby we use Bayesian approach to reliably estimate, under low data regime, independence test statistics used by frequentist independence tests. For both MI estimation (Sec.3.1) and hypothesis likelihood estimation (Sec.3.2), we employ Bayesian inference to calculate statistics by considering the entire parameter space instead of using a point estimate one. Given the estimated Bayesian statistics, we follow the standard frequentist framework to perform independence test. 
The proposed Bayesian-augmented independence tests are then applied to improve the constraint-based causal structure learning. 
We evaluate both local and global causal discovery performance with proposed independence tests on benchmark datasets and compare them to state-of-the-art methods. We empirically demonstrate the effectiveness of the proposed Bayesian approaches in improving both the accuracy and efficiency of the local and global causal discovery under insufficient data. 

\section{Related Work}
To handle causal discovery under insufficient data, some methods downsize the problem domain to sub-domains. 
Rohekar \textit{et al.},~\shortcite{rohekar2020single} approximated the structure by performing independence tests with a small fixed size of the condition set. The structure was then refined by iteratively increasing the condition set. 
A similar idea was explored in the Recursive Autonomy Identification (RAI) method~\cite{yehezkel2009bayesian}.
Related works along this line always assume that there exist sufficient data for sub-domains.  
Besides, Claassen and Heskes~\shortcite{claassen2012bayesian} estimated the posterior distribution of the independence hypothesis between two variables, based on which reliability was quantified. The causal discovery was then processed in decreasing order of reliability. Rohekar \textit{et al.},~\shortcite{rohekar2018bayesian} estimated posterior distribution of DAG through bootstrap samples. The negative effect from independence tests error was minimized through model averaging. 

Some causal discovery methods address the limited data issue by directly improving the independence test~\cite{marx2018stochastic}. 
A Bayesian-augmented frequntist independence test based on Bayes Factor (BF) was proposed~\cite{natori2017consistent} whereby Bayesian parameter estimate is employed in computing BF while the value of BF is then applied to a frequentist independence test.
The proposed independence test is incorporated into RAI, achieving competitive DAG learning performance. However, a threshold is required in~\cite{natori2017consistent} and the selection of threshold can be heuristic. Instead, we propose to formulate the Bayes Factor into a well-defined statistical independence test without requiring threshold tuning.

In addition, different approaches have been proposed for robust independence tests under insufficient data. These methods, however, are not aimed at improving the causal discovery performance. Seok and Seon Kang~\shortcite{article} improved the estimation of mutual information (MI) by partitioning the whole sample space into sub-regions.  
For better MI estimation under limited data, Bayesian approaches have been widely considered~\cite{hutter2002distribution,archer2013bayesian}. 
Besides, shrinkage estimators have also been employed~\cite{sechidis2019efficient,hausser2009entropy}. Another category of recent independence test techniques are focused on developing non-parametric methods to improve efficiency, such as CCIT~\cite{sen2017model} and RCIT~\cite{strobl2019approximate}. These works assume the availability of sufficient data and are mainly focused on continuous variables, while we are focused on discrete ones. 

\section{Proposed Methods}
We consider two types of independence test: MI based and statistical testing based independence tests.
We introduce Bayesian approaches to improve both types of independence tests through a Bayesian-augmented frequentist framework. Particularly, for MI based approach, we employ empirical Bayesian approach for better MI estimation under limited data.  For statistical testing based approach, we consider the empirical Bayesian estimation of hypothesis likelihood and formulate it into $\chi^2$ statistical independence test, providing an accurate $p$-value under limited data.

\subsection{Bayesian Approach for Mutual Information Based Independence Test}
The mutual information (MI) of two discrete random variables $X$ and $Y$ 
is defined as $MI(X;Y) = \sum_{i=1}^{K_x}\sum_{j=1}^{K_y}P(x_i,y_j)\log\frac{P(x_i,y_j)}{P(x_i)P(y_j)}$,
where $K_x$ and $K_y$ denote the total number of possible states of $X$ and $Y$ respectively. $P(x_i,y_j)$,
$P(x_i)$, and $P(y_j)$ 
represent the joint probability of $(X,Y)$, and the marginal probabilities of $X$ and $Y$ respectively.
By definition, 
$MI(X;Y)=0$ if and only if $X$ and $Y$ are independent. In practice, the true MI is unknown, and the estimated MI is always larger than zero. 
In the following, we denote the probability distribution parameters as $\bm\theta$, i.e., $P(x_i)=\theta_i, P(y_j)=\theta_j$ and $P(x_i,y_j)=\theta_{ij}$.
Conventionally, MLE is employed to estimate $\bm{\theta}$ from data as $\hat{\bm{\theta}}= \arg\max_{\bm{\theta}}p(D|\bm{\theta})$,
where $P(D|\bm{\theta})$ is the likelihood of parameter $\bm{\theta}$ given the data $D$.
MI is then estimated as $MI=MI(X;Y|\hat{\bm{\theta}})$. 
When data is insufficient, MLE is not reliable~\cite{geweke1980interpreting} and MI tends to be overestimated. Instead, 
the full Bayesian MI is estimated from data over the entire parameter and hyper-parameter space, i.e.,
\begin{equation}
\resizebox{.9\hsize}{!}{
    $\begin{split}\hat{MI^{fB}} &= MI(X;Y|D)\\
    &=\int\int MI(X;Y|\bm{\theta}, \alpha)p(\bm{\theta}, \alpha|D)d\bm{\theta} d\alpha \\
    &= \int\int MI(X;Y|\bm{\theta})p(\bm{\theta}|\alpha,D)p(\alpha|D)d\bm{\theta} d\alpha\end{split}$}
\label{eqn5}
\end{equation}
where $\alpha$ is the hyper-parameter for symmetric Dirichlet prior of $\bm{\theta}$\footnote{As we have no prior preference on the elements of the Dirichlet distribution, 
we assume symmetric Dirichlet distribution, i.e., each entry in $\bm\alpha$ shares the same value and we denote it as $\alpha$. }. 
The full Bayesian MI is the expected MI over the joint posterior distribution of the parameters and hyper-parameter, i.e., $p(\bm{\theta}, \alpha|D)$. The integration over hyper-parameter $\alpha$ can be computationally challenging~\cite{archer2013bayesian}. Instead of marginalizing out $\alpha$, we propose to maximize it out. Particularly, we approximate the integration over the hyper-parameter space by its mode $\alpha^*$ that maximizes a posterior (MAP) of $\alpha$, i.e., 
$\alpha^* = \arg\max_{\alpha} p(\alpha |D)$. By assuming uniform distribution of $p(\alpha)$, we have $\alpha^* = \arg\max_{\alpha} p(D|\alpha)$. The likelihood $p(D|\alpha)$ can be computed as (See Appx.A for details),
\begin{equation}
\begin{split}
    p(D|\alpha) 
&=N!\frac{\Gamma(K\alpha)}{\Gamma(K\alpha+N)}\prod_{i=1}^K\frac{\Gamma(\alpha+n_i)}{\Gamma(\alpha)n_i!}
\end{split}
\label{eqn7}
\end{equation}
where $K$ is the number of states for the random variable, $n_i$ is the number of samples for state $i$, and $N=\sum_i^Kn_i$.
$P(D|\alpha)$ follows Polya distribution and $\Gamma(x)$ is the gamma function.
We solve for $\alpha^*$ with a fixed-point update~\cite{minka2000estimating}. 

Given $\alpha^*$, the full Bayesian method is converted to the empirical Bayesian method, and we have the proposed empirical Bayesian MI $\hat{MI}^{eB}$ defined as,
\begin{equation}
\label{eqn6}
\hat{MI}^{eB} =\int MI(X;Y|\bm{\theta})p(\bm{\theta}|D, \alpha^*)d\bm{\theta}
\end{equation}
with a closed-form solution (See Appx.A for details):
\begin{equation}
\begin{split}
&\resizebox{\hsize}{!}{$\begin{split}
    \hat{MI}^{eB} &=\psi(N+\alpha^* K+1)-
    \sum_{ij}\frac{n_{ij}+\alpha^*}{N+\alpha^* K}[\psi(n_{i}+\alpha^* K_y+1) \\
    &+\psi(n_{j}+\alpha^* K_x+1) -\psi(n_{ij}+\alpha^*+1)
\end{split} $}
\end{split}
\end{equation}
where $\psi(x)$ is the digamma function.  $n_i$ and $n_j$ are the number of samples for $X=i$ and $Y=j$ respectively, and $n_{ij}$ is the number of samples for $(X,Y)=(i,j)$. The closed-form solution to empirical Bayesian estimation of MI was firstly presented in~\cite{hutter2002distribution}, based on which we propose our approach. Our contribution lies in automatically estimating $\alpha^*$ by maximizing $p(\alpha|D)$ instead of manually selecting $\alpha$ as in Hutter's method.
Given the estimated MI, we compare it against a pre-defined threshold for independence test. If MI is smaller than the threshold, two random variables will be declared to be independent, and dependent otherwise.

\subsection{Bayesian Approach for Statistical Testing Based Independence Test}
We now introduce our proposed Bayesian approach to improve the statistical testing based independence test. We firstly consider a standard independence test, $G$ test~\cite{mcdonald2009handbook}, which is a likelihood ratio test with null hypothesis assuming two random variables are independent. $G$ test is a widely used statistical test. As the same with other statistical tests, G test doesn't require threshold tuning and the significance level is set to be $5\%$ by default.
The formula for the statistic $G$ reads as $G = -2\sum_{i=1}^{K_x}\sum_{j=1}^{K_y} n_{ij}\ln\frac{\hat{\theta}_{i}\hat{\theta}_{j}}{\hat{\theta}_{ij}}$
where $\hat{\hm\theta} = \arg\max_{\bm{\theta}} P(D|\bm{\theta})$. 
Samples $D = \{D_n\}_{n=1}^N$ are $i.i.d$ given parameter $\hat{\bm{\theta}}$ and the statistic $G$
follows asymptotic $\chi^2_{df = (K_x-1)(K_y-1)}$ distribution, based on which a statistical test can be performed (See Appx.B for detailed derivations). As MLE parameter estimates are not reliable under insufficient data, leading to inaccurate estimation of the likelihood of hypothesis, we instead consider the empirical Bayesian estimation. Specifically, we employ the Bayes Factor (BF)~\cite{kass1995bayes} which defines the ratio of expected likelihoods of null hypothesis$(H_0)$ and that of the alternative hypothesis$(H_1)$ over all possible parameter settings  with the posterior distributions of parameters under null and alternate hypothesis respectively, 
\begin{equation}
\label{eqn10}
    BF = \frac{P(D|H_0, \alpha^0)}{P(D|H_1, \alpha^1)} = \frac{\int P(D| \bm{\theta})P(\bm{\theta}|H_0,\alpha^0)d\bm{\theta}}{\int P(D| \bm{\theta})P(\bm{\theta}|H_1,\alpha^1)d\bm{\theta}}
\end{equation}
where $\alpha^0$ and $\alpha^1$ are the hyper-parameters for the symmetric Dirichlet prior under null and alternate hypothesis respectively. 
Both hypothesis likelihoods $P(D|H_0, \alpha^0)$ and $P(D|H_1, \alpha^1)$ can be analytically solved, and BF can be computed (See Appx.C for detailed derivations). However, BF can't be directly applied to a statistical test because samples $D = \{D_n\}_{n=1}^N$ are not $i.i.d$ given hyper-parameter $\alpha$ and $BF$ no longer follows the $\chi^2$ distribution under the null hypothesis. 
Detailed discussions on this are in Appx.C. Instead, we propose to approximate $P(D|\alpha)$ by a multinomial distribution and calculate modified parameters of multinomial distribution with $\alpha$ taken into account, as both capture the distributions for integer random variables, 
i.e., 
\begin{equation}
\label{eqn14}
    P(D|\alpha) \approx P(D|\Tilde{\bm{\theta}}) = \frac{N!}{\prod_{i=1}^Kn_i!}\prod_{i=1}^K\Tilde{\theta_i}^{n_i}
\end{equation}
\noindent where $K$ is the total number of states, and $N=\sum_{i=1}^Kn_i$ is the total number samples with $n_i$ being the number of samples for state $i$. $\Tilde{\theta}_i$ are the modified parameters of the multinomial distribution. $\Tilde{\theta}_i=
\frac{g(n_i, \alpha)}{g(N, K\alpha)}$ with 
$g(n_i, \alpha) = an_i + b\alpha$
where $\Lambda = \begin{pmatrix} a\\b\end{pmatrix}$ are unknown coefficients. In summary, the motivations for the proposed approximation in Eq.~\ref{eqn14} are two folds: 1) we formulate BF into a statistical test whereby threshold 
can be automatically decided; 
2) different from $G$ statistic using MLE parameter estimates, we use modified parameter estimates 
with prior $\alpha$ incorporated.
\noindent By plugging the $P(D|\alpha)$ (defined in Eq.~\ref{eqn7}) into Eq.~\ref{eqn14}, it is clear that to satisfy Eq.~\ref{eqn14}, we must have $ n_i\ln g(n_i, \alpha) = \ln\Gamma(n_i+ \alpha) - \ln\Gamma(\alpha)$.
Given $\{n_i\}_{i=1}^K$ and $\alpha$, we can construct a system of $K$ such equations through which we can solve for $\Lambda^*$, i.e.,
\begin{equation}
    \Lambda^*=\arg\min_{\Lambda}||M\Lambda-T||^2_2 = (M^tM)^{-1}M^tT
\end{equation}
with 
$M = \begin{pmatrix}n_1, \alpha\\n_2, \alpha\\ ... \\ n_K, \alpha\end{pmatrix}$, $T = \begin{pmatrix}t(n_1, \alpha)\\ t(n_2, \alpha)\\...\\t(n_K, \alpha)\end{pmatrix}$, and $t(n_i, \alpha) = \exp(\frac{1}{n_i}(\ln \Gamma(n_i+\alpha) - \ln\Gamma(\alpha)))$. 
Given $\Lambda^*$, we have $\Tilde{\theta}_i$ as $\Tilde{\theta}_i = \frac{g(n_i, \alpha)}{g(N, K\alpha)} = \frac{a^*n_i+ b^*\alpha}{a^*N+b^*K\alpha}$.
$P(D|\Tilde{\bm{\theta}})$ can well approximate $P(D|\alpha)$ (Empirical justifications can be found in Appx.D). Our proposed approximation is different from the method provided in~\cite{minka2000estimating} where the Polya distribution $P(D|\alpha)$ is interpreted as a multinomial distribution with modified counts $\Tilde{n}_{i}$. In addition, our proposed estimation can better approximate the Polya distribution given the symmetric Dirichlet prior (Detailed derivations and empirical evaluation are in Appx.D) compared to~\cite{minka2000estimating}. We then
approximate the hypothesis likelihood under null and alternative hypothesis respectively and obtain a modified Bayes Factor $\Tilde{BF}$
\begin{equation}
     \Tilde{BF} = \frac{P(D|\Tilde{\bm{\theta}}, H_0)}{P(D|\Tilde{\bm{\theta}}, H_1)} = \frac{\prod_{i=1}^{K_x}\Tilde{\theta_i}^{n_i}\prod_{j=1}^{K_y}\Tilde{\theta_j}^{n_j}}{\prod_{i=1, j=1}^{K_xK_y}\Tilde{\theta_{ij}}^{n_{ij}}}
\end{equation}
We obtain the statistic $BF_{chi2}$ for the statistical test as, 
\begin{equation}
     BF_{chi2} = -2\ln\Tilde{BF} = -2\sum_{i=1}^{K_x}\sum_{j=1}^{K_y}n_{ij}\ln\frac{\Tilde{\theta_i}\Tilde{\theta_j}}{\Tilde{\theta_{ij}}}
\end{equation}
The statistic $BF_{chi2}$ asymptomatically follows the $\chi^2_{df = (K_x-1)(K_y-1)}$ distribution (Details are in Appx.D). If $p$-value is smaller than the significance level, we reject the null hypothesis and accept the alternative hypothesis. It is worth noting that $BF$ can be directly applied for a frequentist independence test where a pre-defined threshold $\eta$ is required~\cite{natori2017consistent}. The value of the threshold is unconstrained~\cite{kass1995bayes} making it hard to be properly selected. Instead, our approach only requires a significant level for independence test which is usually set to be $5\%$ by default. Like conventional likelihood ratio test, our method indeed requires the asymptotic assumption. But with the use of Bayesian estimation, our method is less reliant on asymptotic assumption as demonstrated by experiments.
As conventional likelihood ratio test is sound, our method should also be sound. 

It is generally believed that Bayesian approach for parameter estimation is better than MLE under insufficient data~\cite{kruschke2013bayesian}, which motivates our Bayesian-augmented frequentist approaches. 
We theoretically show that Bayesian estimation is always better than MLE for parameter estimation with smaller estimation variance (Details are in Appx.E). Through exhaustive experiments, we further empirically demonstrate the robustness of Bayesian approaches under limited data through improved performance on both independence test and causal discovery. While our discussion focuses on marginal independence test, our methods can be straightforwardly applied to conditional independence test as they share the same mechanism. In fact, when applied to causal discovery, our methods are applied to primarily perform conditional independence tests.

\section{Experiments}
We evaluate both the local and global constraint-based causal discovery performance on benchmark datasets. Our work is to improve independence tests, so as to improve causal discovery under insufficient data. We thus focus our evaluations on constraint-based methods. 
Through exhaustive experiments, we show that our approaches can significantly improve causal discovery performance in terms of both accuracy and efficiency over state-of-the-art methods. Besides, 
we compare proposed independence tests to state-of-the-art independence tests to further show the effectiveness of the proposed methods.

\noindent\textbf{Experiment Settings.} We employ six benchmark datasets\footnote{https://www.bnlearn.com/bnrepository/.} 
that are widely used for causal discovery evaluation: CHILD, INSURANCE, ALARM, HAILFINDER, CHILD3 and CHILD5. Statistical information of datasets are in Appx. F. The causal discovery performance is evaluated in terms of both accuracy and efficiency. For accuracy, we employ the structural hamming distance (SHD)~\cite{Tsamardinos2006}. SHD computes the number of extra and incorrect (missing and reverse) edges in the learned causal structure compared to the ground truth one. For efficiency, we consider the number of conducted independence test. We perform evaluation on a number of small sized datasets.
These small sample sizes are chosen to mimic insufficient data scenario through significantly small number of samples per configuration. For each sample size, we repeat 10 runs and report the averaged performance over 10 runs. In addition, we report standard derivation of SHD.  All the experiments are performed on a laptop with a 2.3 GHz 8-Core Intel Core i9 processor using CPU only (Specific running time can be found in Appx. F). 

\subsection{Local Constraint-based Causal Discovery} 

\begin{table*}[ht!]
  \centering
  \tabcolsep0.1in
   \scalebox{1}{
  \begin{tabular}{|cc|ccc|ccc|}
    \hline
    \hline
   &&\multicolumn{3}{c|}{\underline{SHD}}&\multicolumn{3}{c|}{\underline{\#Independence Test}}\\
    Dataset & Size &$cI^{eB}$ & $cBF_{chi2}$ & CMB &$cI^{eB}$ & $cBF_{chi2}$ & CMB \\ 
    \hline
    \multirow{2}{*}{CHILD} 
    & 100   &2.90$\pm$0.28 &2.65$\pm$0.40 &5.94$\pm$0.65 &1008 &1154 &16869\\
    & 300  &2.61$\pm$0.26 &2.64$\pm$0.59 &6.95$\pm$0.63 &1709 &1926 &14578\\
    & 500  &2.29$\pm$0.31 &2.24$\pm$0.84 &4.52$\pm$0.58 &2524 &4751 &13873\\ 
    &\textbf{MEAN} &2.60 &\textbf{2.51}  &5.80 &\textbf{1747}  &2610  &15107\\
    \hline
    \multirow{2}{*}{INSURANCE} 
    & 100   &3.89$\pm$0.34 & 3.98$\pm$0.39 & 7.18$\pm$0.66 &1261 & 1363 &22168\\
    & 300  &3.47$\pm$0.21 & 3.24$\pm$0.12 & 7.59$\pm$0.57 &1541 & 2977 &18043\\
    & 500  &3.11$\pm$0.21 &2.98$\pm$0.13 & 7.20$\pm$0.67  &1477 &3949 & 14881\\ 
    &\textbf{MEAN} &3.49  &\textbf{3.40}  &7.32 &\textbf{1426}  &2763  &18364\\
    \hline
    \multirow{2}{*}{ALARM} 
     & 100   &2.69$\pm$0.07 &2.39$\pm$0.19 &5.20$\pm$0.71 &1424 &1109 &27492\\
    & 300  &2.50$\pm$0.19 &2.27$\pm$0.15 &4.36$\pm$0.83 &2398 &3885 &14900\\
    & 500  &2.40$\pm$0.11 &2.26$\pm$0.19 &3.53$\pm$0.62 &2807 &4766 &11328\\ 
    &\textbf{MEAN} &2.53  &\textbf{2.31}  &4.36 &\textbf{2210}  &3253  &17907\\
    \hline
    \multirow{2}{*}{HAILFINDER} 
    &&&&&\\[-1em]
    & 500   &3.33$\pm$0.02 & 4.22$\pm$0.04 & 7.90$\pm$0.11 &676 &1923 &183350\\
    & 800  &3.56$\pm$0.01 & 4.49$\pm$0.13 & 7.12$\pm$0.09 &1098 &2145 &169705\\
    & 1000  &3.56$\pm$0.09 & 4.45$\pm$0.08 & 7.10$\pm$0.11 &1924 &2621 &119815\\ 
     &\textbf{MEAN} &\textbf{3.48}  &4.39  &7.37 &\textbf{1233}  &2229  &157620 \\ 
    \hline
    \multirow{2}{*}{CHILD3} 
    &&&&&\\[-1em]
    & 500   &2.46$\pm$0.23 & 2.53$\pm$0.18 &4.72$\pm$0.28 &7168 & 7417 &14789\\
    & 800  &3.01$\pm$0.13 &2.67$\pm$0.11 &3.57$\pm$0.21 &6720 &7802 &9765\\
    & 1000  &2.90$\pm$0.07 &2.57$\pm$0.23 &3.09$\pm$0.19 &8424 &8285 &9516\\ 
    &\textbf{MEAN} &2.79  &\textbf{2.59}  &3.79 &\textbf{7437}  &7835  &11357\\ 
    \hline
    \multirow{2}{*}{CHILD5} 
    &&&&&\\[-1em]
    & 500   &2.87$\pm$0.05 & 2.62$\pm$0.19 & 5.00$\pm$0.15 &5234 & 11126 &16819\\
    & 800  &2.66$\pm$0.21 & 3.02$\pm$0.13 & 5.75$\pm$0.32 & 8236 & 11424 &51967\\
    & 1000  &2.82$\pm$0.23 & 2.99$\pm$0.07 & 4.34$\pm$0.19 &13384 & 9956 & 36888\\ 
     &\textbf{MEAN} & \textbf{2.78} & 2.88 & 5.03 & \textbf{8951} & 10835 & 26322\\ 
    \hline
  \end{tabular}}
  	\caption{Local causal discovery performance under insufficient data}
  \label{tb:accSTMB}
\end{table*}

For the local causal discovery, we employ Causal Markov Blanket (CMB)~\cite{gao2015local}, which is the state-of-the-art method. CMB employs constraint-based approach, which performs conditional independence test using MI to identify the CMB of a target node. We incorporate the proposed independence tests into CMB and compare to the original CMB. We denote $cI^{eB}$ as the CMB with empirical Bayesian MI estimation and $cBF_{chi2}$ as CMB with $BF_{chi2}$ independence test. SHD is $0$ if learned CMB is identical to the ground truth CMB. Details on algorithm settings (e.g., hyper-parameters) are in Appx. F.

From Table~\ref{tb:accSTMB}, we can see that both $cI^{eB}$ and $cBF_{chi2}$ outperform the CMB on all datasets in terms of both accuracy and efficiency under insufficient data. The number of performed independence test reduces dramatically. On ALARM dataset, $cI_{eB}$ only performs 2210 independence tests on average, while CMB requires $17907$ tests on average. The proposed methods improve the accuracy significantly. On INSURANCE dataset, $cBF_{chi2}$ improves the averaged SHD by 3.92 compared to CMB. From the results we can see that, by introducing Bayesian approaches, both the accuracy and the efficiency can be improved. Comparing the performance between the two proposed methods, $cBF_{chi2}$ achieves overall better accuracy, and $cI^{eB}$ is more efficient with the fewest number of independence test on all datasets. 

It is worth noting that the number of independence test increases with reduced samples in CMB, but decreases with the proposed methods. The reason is that under insufficient data, MLE will lead to an overestimated MI. Hence, conventional MI based independence test is likely to declare dependence when data size is small, resulting in a large number of independence test. As the sample size increases, the incorrect dependency declarations will be corrected and the number of independence tests will decrease. On the other hand, our methods are more accurate and show a preference of independence under insufficient data, resulting a small number of performed independence test.

\subsection{Global Constraint-based Causal Discovery}
Majority of global causal discovery algorithms are under causal sufficiency assumption, whereby all random variables are observed in data and there is no latent variable. However, causal sufficiency assumption can be violated since the real data may fail to capture the values for all the variables, leaving some variables to be latent. To address this issue, several recent causal discovery methods~\cite{ramsey2012adjacency,colombo2012learning} have been developed to identify latent common confounders of the observed variables. In our evaluations, we mainly focus on standard algorithms that are under causal sufficiency assumptions. We firstly employ RAI~\cite{yehezkel2009bayesian} as our baseline and compare to two state-of-the-art methods. Then, to demonstrate that our proposed methods can consistently improve causal discovery performance, we consider well-known DAG learning algorithms: PC~\cite{spirtes2000causation} and MMHC~\cite{tsamardinos2006max} as two additional baselines. In the end, we consider the algorithms without causal sufficiency assumption to demonstrate that our proposed methods can be applied to different causal discovery methods, independent of the existence of latent confounders.

\paragraph{Global causal discovery with causal sufficiency assumption.} We employ RAI as our baseline algorithm and incorporate the proposed independence tests. We denote $rI^{eB}$ as the RAI with empirical Bayesian MI estimation, and $rBF_{chi2}$ as RAI with $BF_{chi2}$ independence test. We compare our approaches to two state-of-the-art methods: RAI-BF method~\cite{natori2017consistent} and PC-stable~\cite{colombo2014order}. SC-PC\footnote{https://github.com/honghaoli42/consistent\_pcalg.} can't be performed under insufficient data smoothly, and thus we exclude this method for comparison.  
\begin{table*}[ht!]
  \centering
  \scalebox{0.9}{
  \begin{tabular}{|c c|cccc|cccc|}
    \hline
    \hline
    &&\multicolumn{4}{c|}{\underline{SHD}}&\multicolumn{4}{c|}{\underline{\#Independence Test}}\\
    Dataset &Size &$rI^{eB}$ & $rBF_{chi2}$ & RAI-BF& PC-Stable&$rI^{eB}$ & $rBF_{chi2}$ & RAI-BF& PC-Stable \\ 
    \hline
    \multirow{2}{*}{CHILD}
    & 100  &21.6$\pm$2.1 &24.2$\pm$2.3&30.4$\pm$3.7& 23.8$\pm$1.7 &283 &314&893& 559\\ 
    & 300  &19.9$\pm$2.7 &17.7$\pm$1.8 &23.5$\pm$4.4& 22.6$\pm$1.9 &342 &546 &997& 986 \\ 
    & 500  &17.6$\pm$1.7 &16.0$\pm$2.9 &22.6$\pm$2.4& 24.4$\pm$2.2 &424 &754 &975& 1317\\ 
    &\textbf{MEAN} &19.7 &\textbf{19.3}  &25.5 & 23.6 &\textbf{350}&538 &955& 954 \\
    \hline
    \multirow{2}{*}{INSURANCE} 
    & 100  &48.9$\pm$1.3 &50.1$\pm$2.9  &54.9$\pm$3.6& 52.0 $\pm$1.5 &486 &604  &905& 1217\\ 
    & 300  &47.3$\pm$0.8  &44.5$\pm$2.0 &46.6$\pm$3.2& 50.2$\pm$3.1 &576  &986 &1011& 1250\\
    & 500  &49.5$\pm$1.8  &39.4$\pm$3.0 &47.1$\pm$2.2&50.7$\pm$2.5 &662  &1200 &1120& 2326\\
    &\textbf{MEAN} &48.6  &\textbf{44.7}  &49.5& 51.0 &\textbf{575}&930&1012& 1598\\
    \hline
    \multirow{2}{*}{ALARM} 
    & 100    &44.5$\pm$2.2 &42.7$\pm$2.3 &48.4$\pm$5.8& 45.8 $\pm$4.9 &891 &958 &1591 & 2215\\
    & 300   &40.7$\pm$3.0 &36.1$\pm$4.5 &35.3$\pm$5.4 & 34.6$\pm$2.7 &1158 &1752 &1881& 3398\\
    & 500   &40.0$\pm$3.1  &29.8$\pm$5.1 &29.8$\pm$5.2&36.5 $\pm$5.7 &1433  &2018 &2098& 3992 \\ 
    &\textbf{MEAN} &41.7  &\textbf{36.2}  &37.8 & 39.0 &\textbf{1161}&1576&1857& 3202\\
    \hline
    \multirow{2}{*}{HAILFINDER} 
    &&&&&\\[-1em]
    & 500   &88.0$\pm$2.0 &98.3 $\pm$1.5 & 118.0$\pm$1.0  &91.6$\pm$1.0 &2024&2587 &6171& 3267\\
    & 800  &85.0$\pm$1.7 &106.3 $\pm$2.1&124.7 $\pm$6.7  &99.7$\pm$1.2 &1983 &3726&7847& 3423\\
    & 1000  &92.3$\pm$4.5 &108.3 $\pm$2.3&131.3 $\pm$3.2  &101.8$\pm$2.2 &2638 &3073&16618& 3603\\ 
       &\textbf{MEAN} &\textbf{88.4}  &104.3  &124.7 & 97.7 &\textbf{2215}  &3129 &10212& 3431\\ 
     \hline
    \multirow{2}{*}{CHILD3} 
    & 500   &67.6$\pm$3.2&54.3$\pm$2.6&79.6$\pm$4.9&81.2$\pm$2.8 &2693&3796&5422&4963 \\ 
    & 800   &65.8$\pm$2.5&52.9$\pm$2.8&74.0$\pm$3.7& 79.9$\pm$2.4 &3941&4587&5106& 6026\\
    & 1000   &61.5$\pm$3.8&52.3$\pm$3.9&71.0$\pm$6.5& 81.4$\pm$2.7 &4723&5170&5980& 6846\\
    & \textbf{MEAN}   &65.0&\textbf{53.2}&74.9& 80.8 &\textbf{3786}&4518&5503& 5945\\ \hline
    \multirow{2}{*}{CHILD5} 
    &&&&&\\[-1em]
    & 500   &122.0$\pm$2.6 & 109.3$\pm$5.1 &134.0 $\pm$2.6 & 113.9$\pm$2.4 &6966 &8646 &10038& 10253\\
    & 800  &121.7$\pm$3.8 & 105.3$\pm$4.0 & 132.3$\pm$6.7 &120.1$\pm$2.9 &10249 &10431&9337& 10708\\
    & 1000  &116.3$\pm$2.9 &105.7$\pm$2.5 & 126.3$\pm$7.0 &123.4$\pm$1.7 &10375 &10494&11174& 11070\\ 
     &\textbf{MEAN} &120.0  &\textbf{106.8}  &126.3  & 119.1 &\textbf{9197}  &9857 &11174& 10677\\ 
    \hline
  \end{tabular}}
  	\caption{Global causal discovery performance under insufficient data}
  \label{tb:accglobal}
\end{table*}
SHD is 0 if the learned DAG and the ground truth DAG belong to the same equivalence class. Details on algorithm settings (e.g., hyper-parameters) are in Appx. F.

From Table~\ref{tb:accglobal}, we can see that $rBF_{chi2}$ outperforms RAI-BF and PC-stable on almost all datasets in terms of both accuracy and efficiency. $rI^{eB}$ also achieves overall better accuracy and significantly improves efficiency. For example, on CHILD3, $rBF_{chi2}$ improves the SHD by $21.7$ and $27.6$ compared to RAI-BF and PC-stable. In terms of efficiency, on HAILFINDER,  $rI^{eB}$ only performs 2215 independence tests in average, while RAI-BF requires 10212 tests in average. Comparing between the two proposed methods, $rBF_{chi2}$ achieves better accuracy and $rI^{eB}$ achieves better efficiency. With the proposed methods, the number of independence tests decreases due to the reduced samples for all datasets, which is consistent with the conclusion we have from the local causal discovery. In addition, both RAI-BF and PC-stable show a preference of independence under insufficient data, leading to the decreased number of independence tests with reduced number of samples.

Since $BF_{chi2}$ essentially is only an approximate of original BF, BF with the optimal threshold should outperform $BF_{chi2}$ in principle. However, selecting the optimal threshold for BF can be challenging and incorrect thresholds can lead to inferior causal discovery performance. 
Instead of fixing the threshold of RAI-BF with its default value, we consider the optimal performance of RAI-BF with tuned thresholds for comparison. According to the results (details can be found in Appx.~F), RAI-BF with the optimally tuned threshold at best achieves comparable performance compared to $rBF_{chi2}$ in terms of both accuracy and efficiency, which is expected. While $rBF_{chi2}$ only requires a fixed significance level (5$\%$ by default) without additional tuning process.

To further show that our proposed methods can consistently improve the causal discovery performance, we consider another two widely used DAG learning algorithms: PC~\cite{spirtes2000causation} and MMHC~\cite{tsamardinos2006max}. We incorporate the proposed methods into PC and MMHC for evaluation. From results (details are in Appx.~F), our proposed methods can consistently improve the DAG learning performance, particularly with PC. For example, on ALARM, PC with $BF_{chi2}$ achieves averaged SHD $40.5$, while PC only achieves averaged SHD $58.2$. Overall, $BF_{chi2}$ achieves better accuracy and $I^{eB}$ achieves better efficiency with both PC and MMHC on different datasets. 

\paragraph{Global causal discovery without causal sufficiency assumption.} To demonstrate that our robust independent tests can also be applied to causal discovery without causal sufficiency assumption, we employ the conservative FCI (cFCI) method~\cite{ramsey2012adjacency} as our baseline.  cFCI is considered as the state-of-the-art  causal discovery algorithm that identifies latent confounders. 
\begin{table}[ht!]
  \centering
  \adjustbox{max width=0.45\textwidth}{
  \begin{tabular}{|c|ccc|ccc|}
    \hline
    \hline
    Dataset&\multicolumn{3}{c|}{\underline{SHD}}&\multicolumn{3}{c|}{\underline{\#Independence Test}}\\
    (\textbf{MEAN}) &$cI^{eB}$ & $cBF_{chi2}$ & cFCI&$cI^{eB}$ & $cBF_{chi2}$ & cFCI \\ 
    \hline
    CHILD &49.1 & \textbf{35.1}  & 50.4 &\textbf{109} & 417 & 2289 \\
    INSURANCE &118.7 & \textbf{94.9}  & 121.3 &\textbf{147} & 593 & 4836 \\
    ALARM &105.3 & \textbf{78.2} & 94.7 &\textbf{397} & 902 & 7361 \\
    HAILFINDER &\textbf{153.2} & 220.2 & 339.0 &\textbf{368} & 2024 & 82683 \\
    CHILD3 &204.3 & 135.2 & \textbf{103.6} &\textbf{692} & 2858 & 4009 \\
    CHILD5&250.7 & \textbf{159.0} & 178.8 &\textbf{1145} & 5161 & 7068\\
    \hline
  \end{tabular}}
  	\caption{Global causal discovery performance (with latent confounder) under insufficient data}
  \label{tb:accglobal_fci}
\end{table}
We denote $cI^{eB}$ as the cFCI with empirical Bayesian MI estimation, and $cBF_{chi2}$ as cFCI with $BF_{chi2}$ independence test. We compare our approaches to cFCI with default $g^2$ statistical based independence test\footnote{https://github.com/striantafillou/causal-graphs.}. 
As we can see from Table~\ref{tb:accglobal_fci}, $cI^{eB}$ achieves best efficiency by performing the smallest number of independence tests. In terms of accuracy, $cBF_{chi2}$ achieves overall better performance. The consistent performance improvement further demonstrates that the proposed independence test can improve the causal discovery performance under insufficient data, independent of the existence of latent confounders.

\subsection{Bayesian Approaches for Independence Tests}
To compare the proposed independence tests to state-of-the-art methods, we firstly perform a direct evaluation of proposed independence tests on synthetic data, and we then compare to state-of-the-art methods in terms of causal discovery performance on benchmark datasets. On synthetic data, we compare to three state-of-the-art independence tests: adaptive partition~\cite{article}, empirical Bayesian with fixed $\alpha$~\cite{hutter2002distribution} and full Bayesian method~\cite{archer2013bayesian}. 
We evaluate the performance in terms of both accuracy and efficiency. Experimental results show that the proposed methods achieve better accuracy with significantly improved efficiency. Detailed experiment settings and results can be found in Appx.G. More importantly,  
we compare proposed independence tests to two state-of-the-art methods: adaptive partition and
empirical Bayesian with fixed $\alpha$ methods in terms of causal discovery performance on benchmark datasets. Because the full Bayesian method is of high computational complexity, making it impractical to be applied to constraint-based causal discovery, we exclude the comparison to this method. We incorporate the adaptive partition method and the empirical Bayesian with fixed $\alpha$ method to RAI (denoted as $rI^{AdP}$ and $rI^{eBFix}$ respectively). 
\begin{table}[ht!]
  \centering
  \scalebox{0.8}{
  \begin{tabular}{|c|ccccc|}
    \hline
    \hline
    Dataset&\multicolumn{5}{c|}{\underline{SHD}}\\ 
    (\textbf{MEAN}) &$rI^{AdP}$ & $rI^{eBFix}$ & $rI^{eB}$&$rBF_{chi2}$ & RAI-BF \\ \hline
    CHILD &26.5 & 23.9 & 19.7 & \textbf{19.3} & 25.5 \\
    INSURANCE & 53.2 & 49.1 & 48.6 & \textbf{44.7} & 49.5 \\
    ALARM &46.9 & 40.9 & 41.7& \textbf{36.2} &37.8 \\
    HAILFINDER &\textbf{70.8}&91.2&88.4  &104.3  &124.7\\
    CHILD3 &81.6 & 66.3 & 65.0 & \textbf{53.2} & 74.9\\ 
    CHILD5 &129.9&121.6&120.0  &\textbf{106.8}  &126.3\\\hline
    \end{tabular}}
    \caption{Accuracy comparison to SoTA independence tests}
  \label{tb:CItestsAcc}
\end{table}
As shown in Table~\ref{tb:CItestsAcc}, our methods achieve overall better accuracy than $rI^{AdP}$ and $rI^{eBFix}$ on different datasets. For example, on CHILD3, $rBF_{chi2}$ achieves averaged SHD $53.2$, significantly better than $rI^{AdP}$ which achieves averaged SHD $81.6$. In terms of efficiency evaluation, $rI^{eB}$ also achieves competitive efficiency (details are in Appx. G). 
Overall, our proposed methods outperform other SoTA independence tests in terms of causal discovery performance. On HAILFINDER, because $rI^{AdP}$ tends to declare independence, the learned DAG contains fewer false positive edges compared to other methods and thus its averaged SHD is the best. 

\section{Conclusion}
In this paper, we introduce Bayesian methods for robust constraint-based causal discovery under insufficient data. Two Bayesian-augmented frequentist independence tests are proposed for reliable statistic estimation under a frequentist independence test framework. Specifically, we propose: 1) an effective empirical Bayesian method for accurate estimation of mutual information under limited data; 2) a Bayesian statistical testing method for independence test by formulating the Bayes Factor into the well-defined $\chi^2$ statistical test. 
We apply  the proposed methods to both local and global causal discovery algorithms and evaluate their performance against state-of-the-art methods on different benchmark datasets. The experiments show that, by introducing Bayesian approaches, the proposed methods not only outperform the competing methods in terms of accuracy, but also improve efficiency significantly. 

\section*{Acknowledgments}

This work is supported in part by a DARPA grant FA8750-17-2-0132, and
in part by the Rensselaer-IBM AI Research Collaboration (http://airc.rpi.edu), part of the IBM AI
Horizons Network (http://ibm.biz/AIHorizons).


\section*{Appendix}


\subsection*{A. Full Bayesian MI method}
We define the full Bayesian MI estimation (Eq.~1) as
\begin{equation}
    \begin{split}
        \hat{MI^{fB}} &= \int\int MI(X;Y|\bm{\theta}, \alpha)p(\bm{\theta}, \alpha|D)d\bm{\theta} d\alpha\\
       &= \int\int MI(X;Y|\bm{\theta})p(\bm{\theta}|\alpha)p(\alpha|D)d\bm{\theta} d\alpha
    \end{split}
\end{equation}
To approximately solve for $\hat{MI^{fB}}$, we propose to approximate the integration over hyper-parameters by its mode $\alpha^*$ that maximizes the posterior of $\alpha$ given data $D$, and we obtain the proposed empirical Bayesian MI, 
\begin{equation}
    \begin{split}
        \hat{MI}^{eB} = \int MI(X;Y|\bm{\theta})p(\bm{\theta}|D, \alpha^*)d\bm{\theta} 
    \end{split}
\end{equation}
where $\alpha^* = \arg\max p(\alpha |D)$.
Apply Bayes's rule, we have
\begin{equation}
    p(\alpha|D) \propto p(D|\alpha)p(\alpha)
\end{equation}
By assuming uniform distribution for $p(\alpha)$, we have
\begin{equation}
    \alpha^* = \arg\max p(\alpha |D) = \arg\max p(D |\alpha)
\end{equation}
The likelihood distribution $p(D|\alpha)$ follows the Pólya distribution and is computed as
\begin{equation}
    \begin{split}
        p(D|\alpha) &= \int p(D|\bm{\theta})p(\bm{\theta}|\alpha)d\bm{\theta}
    \end{split}
\end{equation}
where $p(D|\bm{\theta})$ is a mutinomial distribution and $p(\bm{\theta}|\alpha)$ is a Dirichlet distribution. Eq.~14 can be solved analytically
\begin{equation}
    \begin{split}
        p(D|\alpha) &= \int p(D|\theta)p(\bm{\theta}|\alpha)d\bm{\theta}\\
        &=\int \frac{N!}{\prod_{i=1}^{K}n_i!}\prod_{i=1}^K\theta_i^{n_i}\frac{\Gamma(K\alpha)}{\prod_{i=1}^K\Gamma(\alpha)}\prod_{i=1}^K\theta_i^{\alpha-1}d\bm{\theta}\\
        &=\frac{N!}{\prod_{i=1}^{K}n_i!}\int\frac{\Gamma(K\alpha)}{\prod_{i=1}^K\Gamma(\alpha)}\prod_{i=1}^K\theta_i^{n_i+\alpha-1}d\bm{\theta}\\
        &=\frac{N!}{\prod_{i=1}^{K}n_i!}\frac{\Gamma(\alpha K)}{\Gamma(\alpha K+N)}\frac{\prod_{i=1}^K\Gamma(\alpha+n_i)}{\Gamma(\alpha)^K}\\
        &=\frac{N!\Gamma(\alpha K)}{\Gamma(\alpha)^K\Gamma(\alpha K+N)}\prod_{i=1}^{K}\frac{\Gamma(\alpha+n_i)}{n_i!}\\
    \end{split}
\end{equation}
We solve for $\alpha^*$ with a fixed-point update
\begin{equation}
    \alpha^{new} = \alpha^{old}\frac{\sum_{i,j}\psi(\alpha + n_{ij}) - K\psi(\alpha)}{K\psi(\alpha K + N) - \psi(K\alpha)}
\end{equation}
where $\psi$ is the digamma function. Given $\alpha^*$, $\hat{MI}^{eB}$ can be computed as (Eq.~4)
\begin{equation}
    \begin{split}
        &\hat{MI}^{eB} = \int MI(X;Y|\bm{\theta})p(\bm{\theta}|D, \alpha^*)d\bm{\theta} \\ 
        &=\psi(N+\alpha^* K+1)-\sum_{ij}\frac{n_{ij}+\alpha^*}{N+\alpha^* K}[\psi(n_{i}+\alpha^* K_y+1)\\
        &+\psi(n_{j}+\alpha^* K_x+1)- \psi(n_{ij}+\alpha^*+1)]
    \end{split}
\end{equation}
where $\psi$ is the digamma function.

\subsection*{B. Underlying distribution of a Likelihood Ratio Test}
The likelihood ratio test is defined as
\begin{equation}
    LR = -2\ln\frac{\sup_{\bm{\theta}\in\Theta_0}P(D|\bm{\theta})}{\sup_{\bm{\theta}\in\Theta}P(D|\bm{\theta})} 
\end{equation}
where a null hypothesis is corresponding to parameter $\bm{\theta} \in \Theta_0$ and $\Theta_0$ is a subset of the whole parameter space $\Theta$.

\subsubsection*{B.1. Underlying distribution in general}
In this section, we will derive a general form of the underlying distribution for the likelihood ratio test statistic without explicitly providing a likelihood function. The likelihood ratio test statistic can be re-written as
\begin{equation}
    LR = -2(L(\bm{\theta}^0) - L(\bm{\theta}^1))
\end{equation}
where $L(\bm{\theta}) = \ln P(D|\bm{\theta})$ is a log-likelihood function and $\bm{\theta}^0 \in \Theta_0$, $\bm{\theta}^1 \in \Theta$.
We do a Taylor expansion for $L(\bm{\theta}^0)$ at point $\bm{\theta}^1$, and we have
\begin{equation}
\begin{split}
    L(\bm{\theta}^0) &\approx L(\bm{\theta}^1) + (\bm{\theta}^0 - \bm{\theta}^1)L'(\bm{\theta}^1) \\
    &+ \frac{1}{2}(\bm{\theta}^0 - \bm{\theta}^1)^TL''(\bm{\theta}^1)(\bm{\theta}^0 - \bm{\theta}^1)
\end{split}
\end{equation}
Note that $\bm{\theta}^1 = \arg\max P(D|\bm{\theta}, H_1)$ and thus $L'(\bm{\theta}^1) = \bm{0}$. Then we have
\begin{equation}
    L(\bm{\theta}^0) = L(\bm{\theta}^1) + \frac{1}{2}(\bm{\theta}^0 - \bm{\theta}^1)^TL''(\bm{\theta}^1)(\bm{\theta}^0 - \bm{\theta}^1)
\end{equation}
Plug equation(12) into equation(10) and we have
\begin{equation}
    LR =(\bm{\theta}^0 - \bm{\theta}^1)^T(-L''(\bm{\theta}^1))(\bm{\theta}^0 - \bm{\theta}^1)
\end{equation}
In order to derive a general form for the distribution of $LR$ from equation(13), we need to compute the distribution of $(\bm{\theta}^0 - \bm{\theta}^1)$. We do another Taylor expansion for $L'(\bm{\theta}^0)$ at point $\bm{\theta}^1$ and we have
\begin{equation}
    L'(\bm{\theta}^0) = L'(\bm{\theta}^1) + (\bm{\theta}^0 - \bm{\theta}^1)L''(\bm{\theta}^1)
\end{equation}
where $L'(\bm{\theta}^1)=\bm{0}$. And we have
\begin{equation}
    \bm{\theta}^0 - \bm{\theta}^1 = (L''(\bm{\theta}^1))^{-1}L'(\bm{\theta}^0)  
\end{equation}
We plug equation(15) into equation(13), and we have
\begin{equation}
\begin{split}
    LR &=L'(\bm{\theta}^0)^T(L''(\bm{\theta}^1))^{-T}(-L''(\bm{\theta}^1))(L''(\bm{\theta}^1))^{-1}L'(\bm{\theta}^0)\\
    &=L'(\bm{\theta}^0)^T(-L''(\bm{\theta}^1))^{-1}L'(\bm{\theta}^0)
\end{split}
\end{equation}
where $L''(\bm{\theta}^1))^{-T} = L''(\bm{\theta}^1))^{-1}$.
Furthermore, we decompose the positive definite matrix $(-L''(\bm{\theta}^1))^{-1}$ as $(-L''(\bm{\theta}^1))^{-1} = Q^2$. In other words, we denote the decomposition as  
\begin{equation}
    (-L''(\bm{\theta}^1))^{-1} = (-L''(\bm{\theta}^1))^{-1/2}(-L''(\bm{\theta}^1))^{-1/2}
\end{equation}
with $(-L''(\bm{\theta}^1))^{-1/2} = Q$. Then we have
\begin{equation}
\begin{split}
    &LR \\
    &=L'(\bm{\theta}^0)^T(-L''(\bm{\theta}^1))^{-1/2}(-L''(\bm{\theta}^1))^{-1/2}L'(\bm{\theta}^0)\\
    &=L'(\bm{\theta}^0)^T(-L''(\bm{\theta}^1))^{-T/2}(-L''(\bm{\theta}^1))^{-1/2}L'(\bm{\theta}^0)\\
    &=[(-L''(\bm{\theta}^1))^{-1/2}L'(\bm{\theta}^0)]^T[(-L''(\bm{\theta}^1))^{-1/2}L'(\bm{\theta}^0)]\\
    &= Z^TZ
\end{split}
\end{equation}
where $Z = (-L''(\bm{\theta}^1))^{-1/2}L'(\bm{\theta}^0)$. Assume $Z$ follows a Gaussian distribution and we compute its expectation and variance in the following.
\begin{equation}
\begin{split}
    E[Z] &= E[(-L''(\bm{\theta}^1))^{-1/2}L'(\bm{\theta}^0)]\\
    &=(-L''(\bm{\theta}^1))^{-1/2}E[L'(\bm{\theta}^0)]
\end{split}
\end{equation}
where
\begin{equation}
\begin{split}
    E[L'(\bm{\theta}^0)] 
    &= E[\frac{d\log P(D|\bm{\theta}^0)}{d\bm{\theta}}]\\
    &= E[\frac{dP(D|\bm{\theta}^0)}{d\bm{\theta}}\frac{1}{P(D|\bm{\theta}^0)}] \\
    &= \int \frac{dP(D|\bm{\theta}^0)}{d\bm{\theta}}\frac{1}{P(D|\bm{\theta}^0)} P(D|\bm{\theta}^0)dD\\
    &= \frac{d}{d\bm{\theta}}\int p(D|\bm{\theta}^0)dD\\
    &= \frac{d}{d\bm{\theta}} (1) = \bm{0}
\end{split}
\end{equation}
Given the result in Eq.~29, we can show that $E[Z] = \bm{0}$. Now we consider the variance of $Z$. We have 
\begin{equation}
\begin{split}
    Var[Z] &= Var[(-L''(\bm{\theta}^1))^{-1/2}L'(\bm{\theta}^0)]\\
    &=(-L''(\bm{\theta}^1))^{-1}Var[L'(\bm{\theta}^0)]
\end{split}
\end{equation}
where $Var[L'(\bm{\theta}^0)] = E[(L'(\bm{\theta}^0))^2] - E^2[L'(\bm{\theta}^0)] = E[(L'(\bm{\theta}^0))^2]$ because $E[L'(\bm{\theta}^0)]=\bm{0}$. We claim that $E[(L'(\bm{\theta}^0))^2] = E[-L''(\bm{\theta}^0)]$ and we prove this claim in the following:
\begin{equation}
\begin{split}
    &E[-L''(\bm{\theta}^0)] \\
    &= E[-\frac{d^2\log P(D|\bm{\theta}^0)}{d\bm{\theta}^2}]\\
    &= E[-\frac{d}{d\bm{\theta}}(\frac{dP(D|\bm{\theta}^0)}{d\bm{\theta}}\frac{1}{P(D|\bm{\theta}^0)})] \\
    &= E[-\frac{d^2P(D|\bm{\theta}^0)}{d\bm{\theta}^2}\frac{1}{P(D|\bm{\theta}^0)} + (\frac{dP(D|\bm{\theta}^0)}{d\bm{\theta}}\frac{1}{P(D|\bm{\theta}^0)})^2]\\
    &= E[-\frac{d^2P(D|\bm{\theta}^0)}{d\bm{\theta}^2}\frac{1}{P(D|\bm{\theta}^0)}] + E[(\frac{d\log P(D|\bm{\theta}^0)}{d\bm{\theta}})^2]\\
    &= E[(L'(\bm{\theta}^0)^2]\\
\end{split}
\end{equation}
where $E[-\frac{d^2P(D|\bm{\theta}^0)}{d\bm{\theta}^2}\frac{1}{P(D|\bm{\theta}^0)}] = \bm{0}$ because
\begin{equation}
\begin{split}
    &E[-\frac{d^2P(D|\bm{\theta}^0)}{d\bm{\theta}^2}\frac{1}{P(D|\bm{\theta}^0)}] \\
    &= -\int \frac{d^2P(D|\bm{\theta}^0)}{d\bm{\theta}^2}\frac{1}{P(D|\bm{\theta}^0)} P(D|\bm{\theta}^0)dD\\
    &= -\frac{d^2}{d\bm{\theta}^2}\int p(D|\bm{\theta}^0)dD\\
    &= -\frac{d^2}{d\bm{\theta}^2} (1) = \bm{0}\\
\end{split}
\end{equation}
After we prove the claim, we plug the claim into Eq.~30 and we have 
\begin{equation}
\begin{split}
    Var[Z] &=(-L''(\bm{\theta}^1))^{-1}E[-L''(\bm{\theta}^0)]
\end{split}
\end{equation}
In the end, we derive a general form for the distribution of $LR$ as
\begin{equation}
    LR = Z^TZ
\end{equation}
where $Z\sim N(\bm{0}, Var[Z])$ with $Var[Z]=(-L''(\bm{\theta}^1))^{-1}E[-L''(\bm{\theta}^0)]$.

\subsubsection*{B.2. G-test and $\chi^2$ distribution}
The value of the G test is derived from the $LR$ where the underlying model is a multinomial model. Suppose we have $i.i.d$ samples $D = \{D_n\} = \{(X_n, Y_n)\}$ where $n = \{1,2,...,N\}$, given the multinomial model, we have
\begin{equation}
\begin{split}
    g = -2\ln\frac{\prod_{n=1}^NP(X_n|\bm{\theta}^0)P(Y_n|\bm{\theta}^0)}{\prod_{n=1}^NP((X_n,Y_n)|\bm{\theta}^1)}
\end{split}
\end{equation}
where we apply the null hypothesis that $P(X,Y) = P(X)P(Y)$. Suppose we have $X = x_i$ where $i = \{1,2,...,K_x\}$ and $Y = y_j$ where $j = \{1,2,...,K_y\}$, then we have
\begin{equation}
\begin{split}
    g = -2\ln\frac{\prod_{i=1}^{K_x}\theta_{i}^{n_{i}}\prod_{i=1}^{K_x}\theta_{j}^{n_{i}}}{\prod_{i=1}^{K_x}\prod_{j=1}^{K_y}\theta_{ij}^{n_{ij}}}
\end{split}
\end{equation}
Note that 
\begin{equation}
\prod_{i=1}^{K_x}\theta_{i}^{n_{i}} = \prod_{i=1}^{K_x}\theta_{i}^{\sum_jn_{ij}} = \prod_{i=1}^{K_x}\prod_{j=1}^{K_y}\theta_{i}^{n_{ij}}
\end{equation}
Similarly, we have $\prod_{j=1}^{K_y}\theta_{j}^{n_{j}} = \prod_{j=1}^{K_y}\prod_{i=1}^{K_x}\theta_{j}^{n_{ij}}$. Then we can write Eq.~36 as
\begin{equation}
\begin{split}
    g &= -2\ln\frac{\prod_{i=1}^{K_x}\prod_{j=1}^{K_y}\theta_{i}^{n_{ij}}\theta_{i}^{n_{ij}}}{\prod_{i=1}^{K_x}\prod_{j=1}^{K_y}\theta_{ij}^{n_{ij}}}\\
    &= -2\sum_{i=1}^{K_x}\sum_{j=1}^{K_y} n_{ij}\ln\frac{\theta_{i}\theta_{j}}{\theta_{ij}}
\end{split}
\end{equation}
which is the definition of the G statistic. Now we show that given a multinomial model which is applied in the G test, $Var[Z] = (-L''(\bm{\theta}^1))^{-1}E[-L''(\bm{\theta}^0)] = 1$ and thus $g$ follows $\chi^2$ distribution. Firstly, given a multinomial model, we have $L(\bm{\theta}) = \log P(D|\alpha) = \log \prod_{n=1}^NP(D_n|\alpha) = \sum_{n=1}^N\log P(D_n|\bm{\theta})$. Then we have
\begin{equation}
\begin{split}
    E[-L''(\bm{\theta}^0)] &= E[-\sum_{n=1}^N\frac{d^2\log P(D_n|\bm{\theta}^0)}{d\bm{\theta}^2}] \\
    &= \sum_{n=1}^N E[-\frac{d^2\log P(D_n|\bm{\theta}^0)}{d\bm{\theta}^2}]
\end{split}
\end{equation}
where $E[-\frac{d^2\log P(D_n|\bm{\theta}^0)}{d\bm{\theta}^2}]$ is a function of the true underlying $\bm{\theta}$ and we denote it as $I_{\bm{\theta}}$ (which is known as the fisher information). We thus have $E[-L''(\bm{\theta}^0)] = N I_{\bm{\theta}}$. Then for the $Var[Z]$, we have
\begin{equation}
\begin{split}
    Var[Z] &= (-L''(\bm{\theta}^1))^{-1}E[-L''(\bm{\theta}^0)] \\
    &= (-L''(\bm{\theta}^1))^{-1}NI_{\bm{\theta}}\\
    &= (\frac{-L''(\bm{\theta}^1)}{N})^{-1}I_{\bm{\theta}}\\
    &= (\frac{1}{N}\sum_{n=1}^N-\frac{d^2\log P(D_n|\bm{\theta}^1)}{d\bm{\theta}^2})^{-1}I_{\bm{\theta}}\\
    &= I_{\bm{\theta}}^{-1}I_{\bm{\theta}} = \mathbb{1}
\end{split}
\end{equation}
where we assume $N$ is large enough so that the averaged $-\frac{d^2\log P(D_n|\bm{\theta}^1)}{d\bm{\theta}^2}$ can well approximate the expectation $I_{\bm{\theta}}$. Then we can show that $g = Z^TZ \sim \chi^2$ with $Z \sim N(0, \mathbb{1})$.

\subsection*{C. Bayes Factor and $\chi^2$ Distribution}
\subsubsection*{C.1. Closed-form solution of Bayes Factor}
We denote the Bayesian likelihood ratio value as $BF$ with the definition (Eq.~5),
\begin{equation}
    BF = \frac{P(D|H_0, \alpha^0)}{P(D|H_1, \alpha^1)} = \frac{\int P(D| \bm{\theta})P(\bm{\theta}|H_0,\alpha^0)d\bm{\theta}}{\int P(D| \bm{\theta})P(\bm{\theta}|H_1,\alpha^1)d\bm{\theta}}
\end{equation}
where we assume the symmetric Dirichlet prior. Under the null hypothesis $H_0$, i.e., two random variables are independent, we set the posterior for each random variable $X,Y$ separately, i.e.,
\begin{equation}
    p(\bm{\theta}|H_0,\alpha^0) = p(\bm{\theta}_x|H_0,\alpha_x)p(\bm{\theta}_y|H_0,\alpha_y)
\end{equation}
with $p(\bm{\theta}_x|H_0,\alpha_x) \sim Dir(\alpha_x)$ and $p(\bm{\theta}_y|H_0, \alpha_y) \sim Dir(\alpha_y)$.
Given parameters under null hypothesis, we have the multinomial distribution for $p(D|\bm{\theta})$,
\begin{equation}
    \begin{split}
        p(D|\bm{\theta}) = N!\prod_{i=1}^{K_x}\prod_{j=1}^{K_y}\frac{(\theta_{x_i}\theta_{y_j})^{n_{ij}}}{n_{ij}!}
    \end{split}
\end{equation}
Then we have $p(D|H_0, \alpha^0)$ as
\begin{equation}
    \begin{split}
        &p(D|H_0, \alpha^0) = \int p(D|\bm{\theta})p(\bm{\theta}|H_0, \alpha^0)d\bm{\theta}\\
        &=\int N!\prod_{i=1}^{K_x}\prod_{j=1}^{K_y}\frac{(\theta_{x_i}\theta_{y_j})^{n_{ij}}}{n_{ij}!}p(\bm{\theta}_x|H_0,\alpha_x)p(\bm{\theta}_y|H_0, \alpha_y)d\bm{\theta}\\
        &=\frac{N!}{\prod_{i=1}^{K_x}\prod_{j=1}^{K_y}n_{ij}!}\\
        &\times\int \prod_{i=1}^{K_x}\prod_{j=1}^{K_y}(\theta_{x_i}\theta_{y_j})^{n_{ij}}p(\bm{\theta}_x|H_0, \alpha_x)p(\bm{\theta}_y|H_0, \alpha_y)d\bm{\theta}\\
    \end{split}
\end{equation}
Notice that
\begin{equation}
    \begin{split}
        \prod_{i=1}^{K_x}\prod_{j=1}^{K_y}(\theta_{x_i}\theta_{y_j})^{n_{ij}} &= \prod_{i=1}^{K_x}\theta_{x_i}^{\sum_jn_{ij}}\prod_{j=1}^{K_y}\theta_{y_j}^{\sum_in_{ij}}\\
        &=\prod_{i=1}^{K_x}\theta_{x_i}^{n_{i}}\prod_{j=1}^{K_y}\theta_{y_j}^{n_{j}}
    \end{split}
\end{equation}
Plug Eq.~45 back into Eq.~44 and we have
\begin{equation}
\begin{split}
    &p(D|H_0, \alpha^0) \\
    &= M\int \prod_{i=1}^{K_x}\theta_{x_i}^{n_{i}}\prod_{j=1}^{K_y}\theta_{y_j}^{n_{j}}p(\bm{\theta}_x|H_0, \alpha_x)p(\bm{\theta}_y|H_0, \alpha_y)d\bm{\theta}\\ 
    &=M\int \prod_{i=1}^{K_x}\theta_{x_i}^{n_{i}}p(\bm{\theta}_x|H_0, \alpha_x)d\bm{\theta}_x\int\prod_{j=1}^{K_y}\theta_{y_j}^{n_{j}}p(\bm{\theta}_y|H_0, \alpha_y)d\bm{\theta}_y\\
\end{split}
\end{equation}
where $M = \frac{N!}{\prod_{i=1}^{K_x}\prod_{j=1}^{K_y}n_{ij}!}$. As the integration over $\bm{\theta}_x$ and $\bm{\theta}_y$ in Eq.~46 are identical, we only derive for $\bm{\theta}_x$,
\begin{equation}
    \begin{split}
        &\int \prod_{i=1}^{K_x}\theta_{x_i}^{n_{i}}p(\bm{\theta}_x|H_0, \alpha_x)d\bm{\theta}_x\\
        &=\int \prod_{i=1}^{K_x}\theta_{x_i}^{n_{i}}\frac{\Gamma(K_x\alpha_{x})}{\prod_{i=1}^{K_x}\Gamma(\alpha_x)}\prod_{i=1}^{K_x}\theta_{x_i}^{\alpha_{x}-1}d\bm{\theta}_x\\
        &=\frac{\Gamma(K_x\alpha_x)}{\Gamma(K_x\alpha_x + N)}\prod_{i=1}^{K_x}\frac{\Gamma(\alpha_{x} + n_{i})}{\Gamma(\alpha_{x})}\\
    \end{split}
\end{equation}
The integration over $\bm{\theta}_y$ can be done in the same way,
\begin{equation}
    \int \prod_{j=1}^{K_y}\theta_{y_j}^{n_{j}}p(\bm{\theta}_y|\alpha_y)d\bm{\theta}_y = \frac{\Gamma(K_y\alpha_y)}{\Gamma(K_y\alpha_y + N)}\prod_{j=1}^{K_y}\frac{\Gamma(\alpha_{y} + n_{j})}{\Gamma(\alpha_{y})}
\end{equation}
Given Eq.~47 and Eq.~48, we have
\begin{equation}
    \begin{split}
        &p(D|H_0, \alpha^0) = \int p(D|\bm{\theta})p(\bm{\theta}|H_0, \alpha^0)d\bm{\theta}\\
        &=\frac{M\Gamma(K_x\alpha_x)\Gamma(K_y\alpha_y)}{\Gamma(K_x\alpha_x + N)\Gamma(K_y\alpha_y + N)}\\
        &\times\prod_{i=1}^{K_x}\frac{\Gamma(\alpha_{x} + n_{i})}{\Gamma(\alpha_{x})}\prod_{j=1}^{K_y}\frac{\Gamma(\alpha_{y} + n_{j})}{\Gamma(\alpha_{y})}\\
    \end{split}
\end{equation}
where $M = \frac{N!}{\prod_{i=1}^{K_x}\prod_{j=1}^{K_y}n_{ij}!}$.
Under the alternative hypothesis, i.e., two random variables are dependent, we set posterior for the joint distribution of two random variables, i.e.,
\begin{equation}
    \begin{split}
        &p(\bm{\theta}|H_1, \alpha^1) \sim Dir(\alpha_{xy})\\
    \end{split}
\end{equation}
with dimension $K = K_xK_y$. Following the similar procedure as we did for the null hypothesis, we can show that
\begin{equation}
    \begin{split}
        &p(D|H_1, \alpha^1) = \int p(D|\bm{\theta})p(\bm{\theta}|H_1, \alpha_{xy})d\bm{\theta}\\
        &=M\frac{\Gamma(\alpha_{xy}K)}{\Gamma(\alpha_{xy}K + N)}\prod_{i}^{K}\frac{\Gamma(\alpha_{xy} + n_{i})}{\Gamma(\alpha_{xy})}\\
    \end{split}
\end{equation}
where $M = \frac{N!}{\prod_{i=1}^{K}n_{i}!}$. Given the marginal likelihood for the null hypothesis (Eq.~49) and and the marginal likelihood for the alternative hypothesis (Eq.~51)), we calculate the likelihood ratio as
\begin{equation}
\begin{split}
    BF = &\frac{\Gamma(\alpha_xK_x)\Gamma(\alpha_y K_y)\Gamma(\alpha_{xy} K + N)}{\Gamma(\alpha_{xy} K)\Gamma(\alpha_x K_x + N)\Gamma(\alpha_y K_y + N)}\\
    &\times\frac{\Gamma(\alpha_{xy})^K}{\Gamma(\alpha_{x})^{K_x}\Gamma(\alpha_{y})^{K_y}} \\
&\times\frac{\prod_{i=1}^{K_x}\Gamma(\alpha_{x} + n_{i})\prod_{j=1}^{K_y}\Gamma(\alpha_{y} + n_{j})}{\prod_{i=1}^{K_x}\prod_{j=1}^{K_y}\Gamma(\alpha_{xy} + n_{ij})}
\end{split}
\end{equation}
Given a pre-defined threshold $\eta$, if $BF>\eta$, the null hypothesis is more likely to be supported by the data and two variables are declared to be independent. Otherwise, we accept alternative hypothesis and declare the variables to be dependent.

\subsubsection*{C.2. BF and $\chi^2$ distribution}
The Bayes factor is a likelihood ratio of the marginal likelihood of two competing hypothesis, usually the null and alternative. The Bayes factor is defined as
\begin{equation}
    BF = \frac{P(D|H_0, \alpha^0)}{P(D|H_1, \alpha^1)} = \frac{\int P(D| \bm{\theta})P(\bm{\theta}|H_0,\alpha^0)d\bm{\theta}}{\int P(D| \bm{\theta})P(\bm{\theta}|H_1,\alpha^1)d\bm{\theta}}
\end{equation}
Different from the G-test which calculates statistics based on one set of parameters, the Bayes factor is considering all possible sets of parameters given the hypothesis.
Note that $P(D|\alpha)$ is a likelihood function with $\alpha = \arg\max P(D|\alpha)$, and we can apply the general form of the underlying distribution of the likelihood ratio test to $BF$. We modify the value $BF$ as
\begin{equation}
    LR_{BF} = -2\ln{BF} = -2\ln \frac{P(D|H_0,\alpha^0)}{P(D|H_1,\alpha^0)} = Z^TZ
\end{equation}
where $Z = (-L''(\alpha^1))^{-1/2}L'(\alpha^0)$. And $Z\sim N(0, Var[Z])$ with $Var[Z]=(-L''(\alpha^1))^{-1}E[-L''(\alpha^0)]$. We can analytically compute the likelihood function by integrating out parameter $\bm{\theta}$ as
\begin{equation}
\begin{split}
    P(D|\alpha) &= \int P(D| \bm{\theta})P(\bm{\theta}|\alpha)d\bm{\theta} \\
    &= \frac{\Gamma(\alpha_0)}{\Gamma(\alpha_0 + N)}\prod_{k}^{K}\frac{\Gamma(\alpha_k + n_{k})}{\Gamma(\alpha_k)}
\end{split}
\end{equation}
where $\alpha_0 = \sum_{k=1}^K\alpha_k$ and $P(D|\alpha)$ is the probability mass function for the Polya distribution. And the log-likelihood function is
\begin{equation}
\begin{split}
    L(\alpha) &= \ln P(D|\alpha)\\
    &= \ln\Gamma(\alpha_0) - \ln\Gamma(\alpha_0 + N) \\
    &+ \sum_{k}^{K}\ln\Gamma(\alpha_k + n_{k}) - \ln\Gamma(\alpha_k)
\end{split}
\end{equation}
In this case, we should treat $D$ as one sample data and thus we can't expect $-L''(\alpha^1)$ approach to $E[-L''(\alpha^0)]$. In other words, we can't naturally have $Var[Z] = \mathbb{1}$ and thus $\Tilde{BF}$ doesn't follow $\chi^2$ distribution. 

\subsection*{D. Bayesian G statistic and its distribution}
\subsubsection*{D.1. An approximated Polya Distribution}
We denote the proposed likelihood ratio value as $BF$ with the definition (Eq.~5),
\begin{equation}
    BF = \frac{P(D|H_0, \alpha^0)}{P(D|H_1, \alpha^1)} = \frac{\int P(D| \bm{\theta})P(\bm{\theta}|H_0,\alpha^0)d\bm{\theta}}{\int P(D| \bm{\theta})P(\bm{\theta}|H_1,\alpha^1)d\bm{\theta}}
\end{equation}
where we assume the symmetric Dirichlet prior. To better decide the threshold, we propose to combine the Bayes Factor $BF$ with a well-defined statistical test $\chi^2$ via an approximated Pólya distribution, i.e.,
\begin{equation}
    P(D|\alpha) \approx P(D|\Tilde{\bm{\theta}}) = \frac{N!}{\prod_{i=1}^Kn_i!}\prod_{k=1}^K\Tilde{\theta_k}^{n_k}
\end{equation}
Because $\frac{N!}{\prod_{i=1}^Kn_i!}$ is a common component that exists in both $P(D|\alpha)$ and $P(D|\Tilde{\bm{\theta}})$, we ignore this term in the following derivations. We re-write the log-likelihood function $\ln P(D|\alpha)$ as
\begin{equation}
\begin{split}
    \ln P(D|\alpha) &= \sum_{k}^K(\ln\Gamma(n_k+ \alpha) - \ln\Gamma(\alpha)) \\
    &- (\ln\Gamma(N + K\alpha) - \ln\Gamma(K\alpha))
\end{split}
\end{equation}
On the other hand, we have the log-likelihood of $P(D|\Tilde{\bm{\theta}})$ as
\begin{equation}
    \ln P(D|\Tilde{\bm{\theta}}) = \sum_k^Kn_k\ln\Tilde{\theta}_k
\end{equation}
Furthermore, we assume that $
\Tilde{\theta}_k$ has the form $\frac{g(n_k, \alpha)}{g(N, K\alpha)}$ where $g(n_k, \alpha)$ is a function with unknown parameters that need to be estimated. We plug $\Tilde{\theta}_k = \frac{g(n_k, \alpha)}{g(N, K\alpha)}$ into Eq.~60 and we have
\begin{equation}
\begin{split}
    \ln P(D|\Tilde{\bm{\theta}}) = \sum_k^Kn_k\ln g(n_k, \alpha) - N\ln g(N, K\alpha)
\end{split}
\end{equation}
Comparing Eq.~59 and Eq.~60, we can see that to make $P(D|\alpha) \approx P(D|\Tilde{\bm{\theta}})$, the desired property of the function $g(n_k, \alpha)$ is
\begin{equation}
    n_k\ln g(n_k, \alpha) = \ln\Gamma(n_k+ \alpha) - \ln\Gamma(\alpha)
\end{equation}
Based on the fact that $\sum_k^K\Tilde{\theta_k} = 1$, the
function $g(n_k, \alpha)$ should subject to the constraint that
\begin{equation}
    \sum_k^Kg(n_k, \alpha) = g(N, K\alpha) =g(\sum_k^Kn_k, \sum_k^K\alpha)
\end{equation}
In other words, the function $g(n_k, \alpha)$ is a linear function with respect to both $n_k$ and $\alpha$. Thus, the form of the function $g(n_k, \alpha)$ should be
\begin{equation}
    g(n_k, \alpha) = an_k + b\alpha
\end{equation}
where parameters $\Lambda = \begin{pmatrix} a\\b\end{pmatrix}$ are unknowns that need to be computed. To estimate two parameters, we have 
\begin{equation}
    \begin{pmatrix}n_1, \alpha\\n_2, \alpha\\ ... \\ n_K, \alpha\end{pmatrix}\begin{pmatrix} a\\b\end{pmatrix} = \begin{pmatrix} t(n_1, \alpha)\\t(n_2, \alpha)\\ ... \\t(n_K, \alpha)\end{pmatrix}
\end{equation}
where $t(n_k, \alpha) = \exp(\frac{1}{n_k}(\ln\Gamma(n_k+ \alpha) - \ln\Gamma(\alpha)))$. Denote $M = \begin{pmatrix}n_1, \alpha\\n_2, \alpha\\ ... \\ n_K, \alpha\end{pmatrix}$,  $T = \begin{pmatrix} t(n_1, \alpha)\\t(n_2, \alpha)\\ ... \\t(n_K, \alpha)\end{pmatrix}$, and we re-write Eq.~65 as $M\Lambda = T$. We always have $K \geq 2$ and we solve for $\Lambda$ with the least square error, i.e.,
\begin{equation}
    \Lambda^* = \arg\min||M\Lambda-T||^2
\end{equation}
with the solution $\Lambda^* = (M^TM)^{-1}M^TT$. Given $\Lambda^*$, we have $\Tilde{\theta}_k$ as
\begin{equation}
    \Tilde{\theta}_k = \frac{g(n_k, \alpha)}{g(N, K\alpha)} = \frac{a^*n_k+ b^*\alpha}{a^*N+b^*K\alpha}
\end{equation}
and $P(D|\Tilde{\bm{\theta}})$ can well approximate $P(D|\alpha)$.

We demonstrate the effectiveness of approximating the Polya probability $p(D|\alpha)$ with modified parameter $\Tilde{\bm{\theta}}$ by comparing it with Minka's approach which approximates the Polya distribution by modified counts. Synthetic data is generated following the procedure stated in the paper. In particular, we consider independency and dependency separately and synthetic data is generated without assuming symmetric Dirichlet prior. 
The relative absolute error between the estimated probability $\Tilde{p}$ and the true probability $p$, i.e., $\frac{|p-\Tilde{p}|}{p}$ is applied as the measurement given each sample set $D$.
We report the average error over 1000 runs for each sample size. 
\begin{figure}[!ht]
   \centering
   \subfloat[Independent Case]{
   \includegraphics[width= 2.3in]{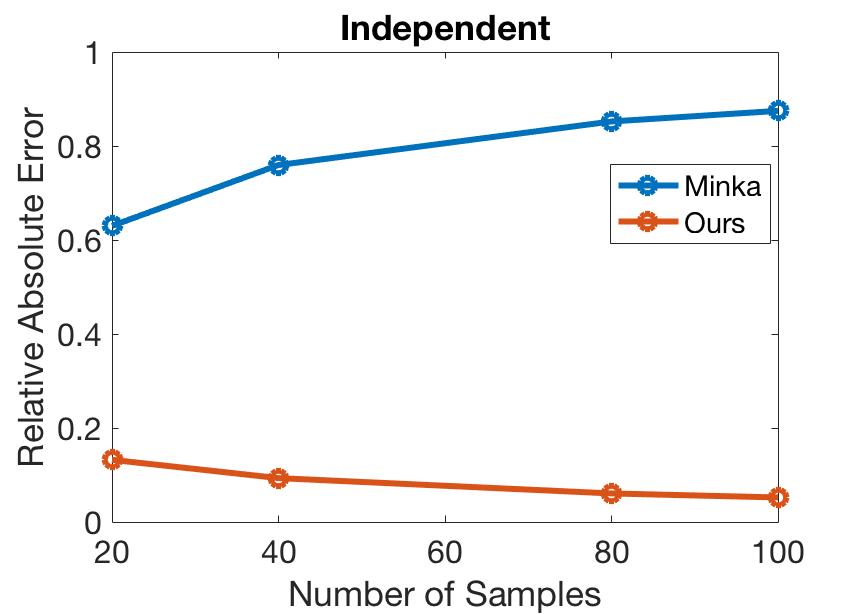}}\\
   \subfloat[Dependent]{\includegraphics[width= 2.3in]{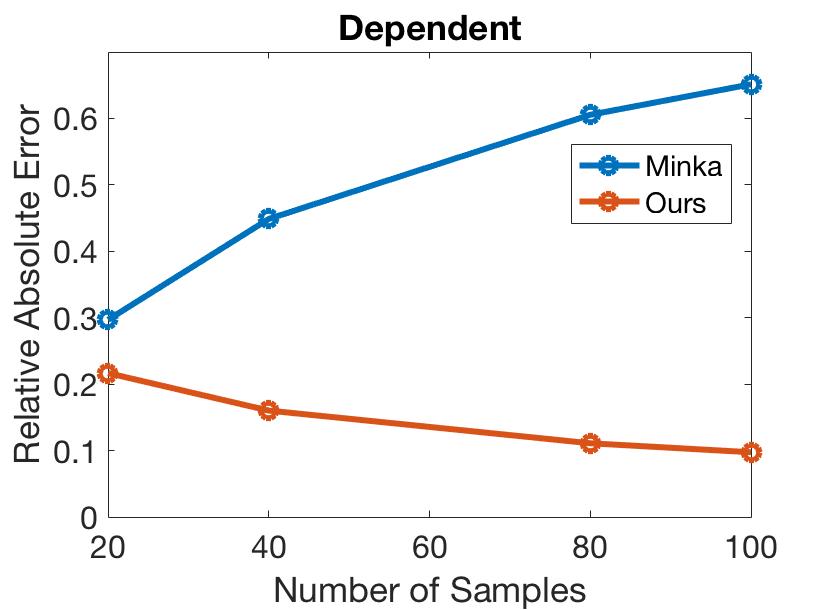}}\\
   \caption{Relative Absolute Error of Polya distribution}
\end{figure}
As we can see from Figure 1, our approach approximates the true polya probability much better than Minka's approach. The reason is that Minka's approach requires the estimation of Dirichlet hyper-parameters and can't work well with symmetric Dirichlet prior. In addition, the visualization of the polya distribution is shown in Figure 2 and Figure 3. 
\begin{figure}[!ht]
   \centering
   \includegraphics[width= 3.3in,height=1.5in]{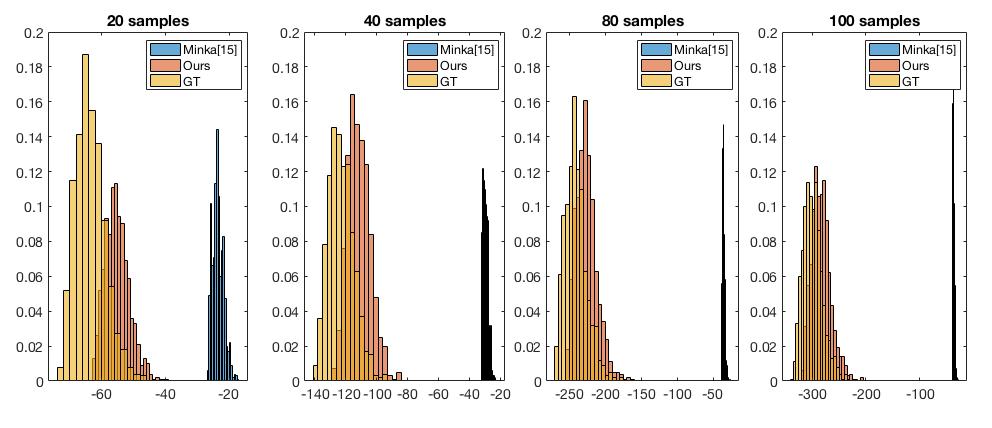}
   \caption{Visualization of Polya distribution Estimation (Independent Case)}
\end{figure}
\begin{figure}[!ht]
   \centering
   \includegraphics[width= 3.3in, height=1.5in]{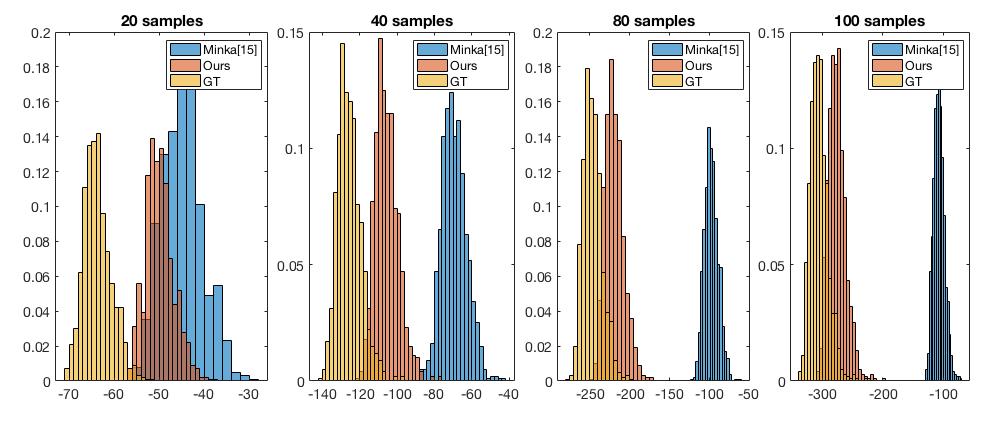}
   \caption{Visualization of Polya distribution Estimation (Dependent Case)}
\end{figure}

\subsubsection*{D.2. Distribution of Bayesian G statistic}
\begin{figure}[ht!]
    \centering
    \includegraphics[width= 3.2in,height= 1.6in]{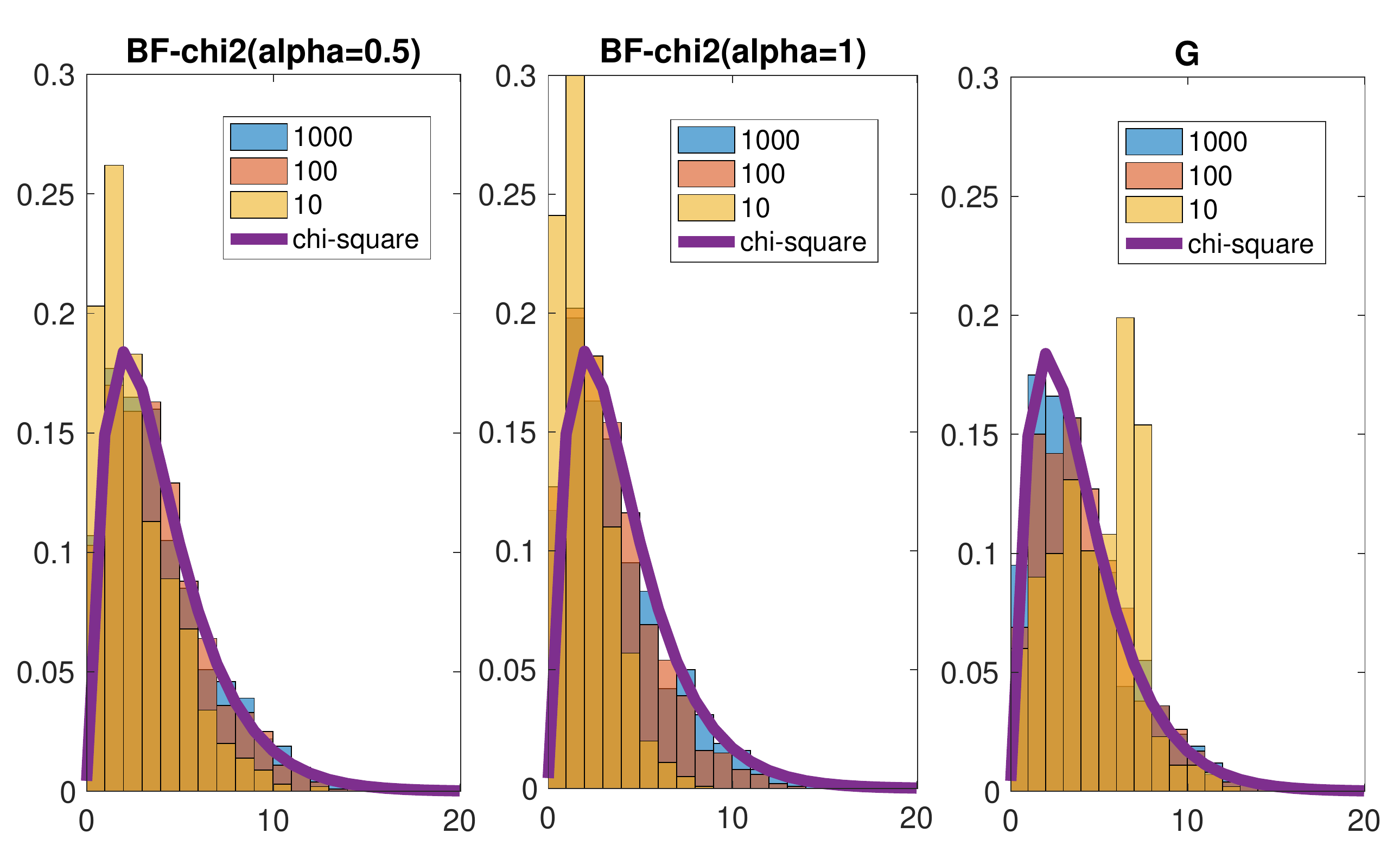}
    \caption{Distribution of the Bayesian G statistic (3 by 3)}
    \label{fig:distribution-3by3}
\end{figure}

\begin{figure}[ht!]
    \centering
    \includegraphics[width= 3.2in,height= 1.6in]{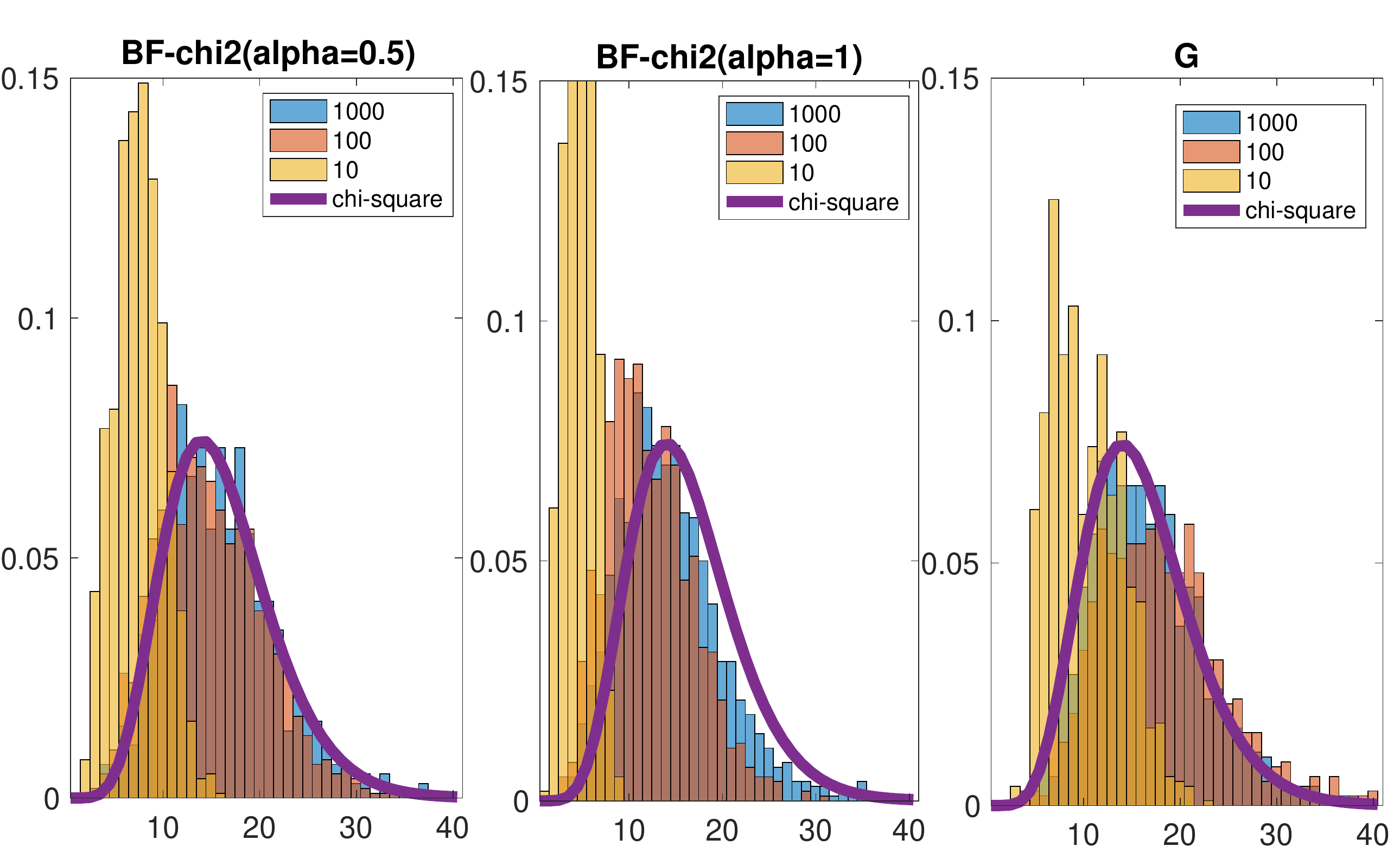}
    \caption{Distribution of the Bayesian G statistic (5 by 5)}
    \label{fig:distribution-5by5}
\end{figure}

\begin{figure}[ht!]
    \centering
    \includegraphics[width= 3.2in,height= 1.6in]{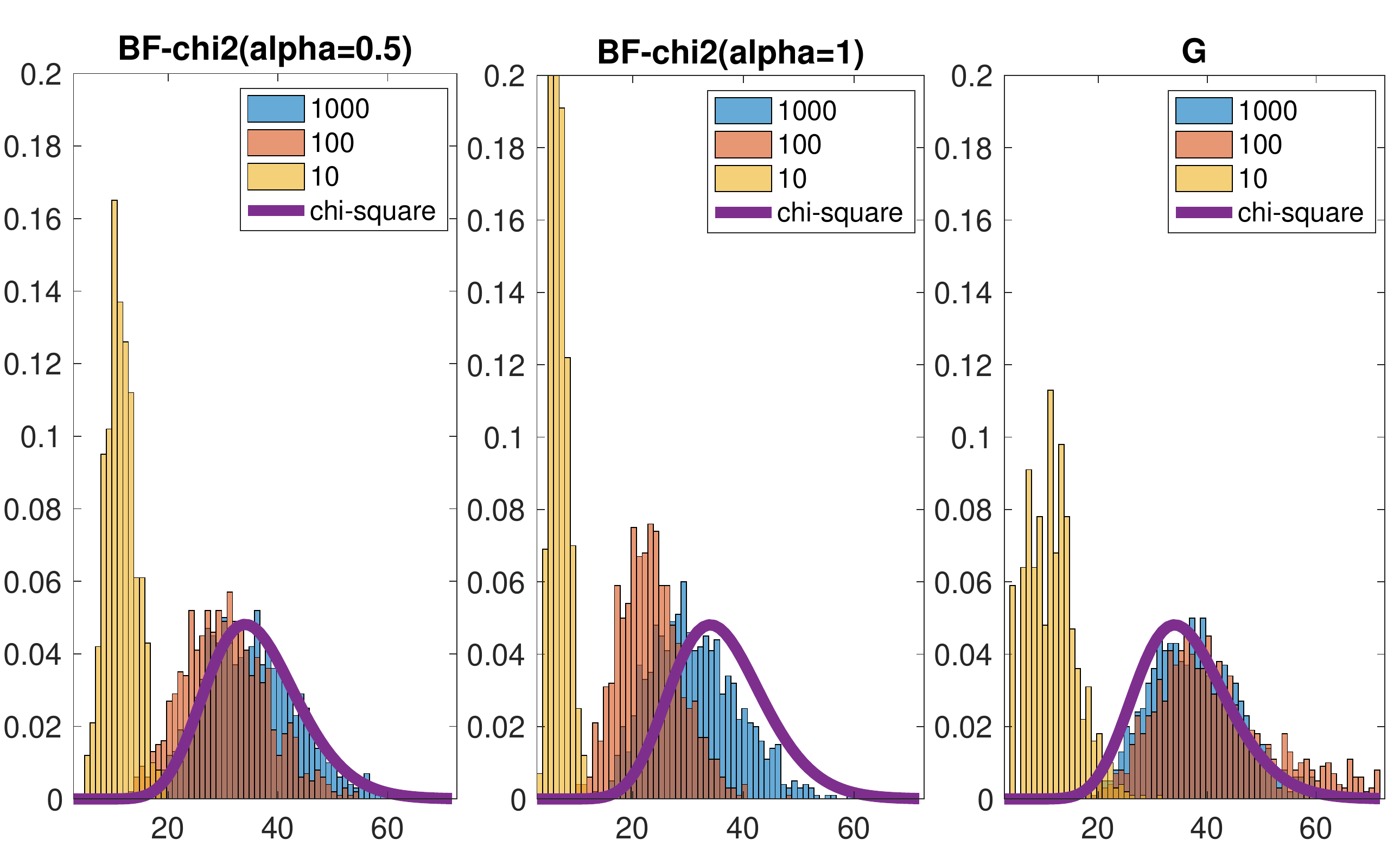}
    \caption{Distribution of the Bayesian G statistic (7 by 7)}
    \label{fig:distribution-7by7}
\end{figure}

We study the distribution of the proposed Bayes Factor statistic $BF_{chi2}$ to show that it asymptomatically follows the $\chi^2$ distribution given the null hypothesis being true, i.e., two variables are independent. We perform experiments on the synthetic data and follow the procedure stated in the paper to generate the synthetic data. We consider both the uniform prior $\alpha=1$ and Jeffrey's prior $\alpha=0.5$ to study the effect of the Dirichlet hyper-parameter. As we estimate the distribution under the true null hypothesis, we set two random variables $X$ and $Y$ to be independent. 
We estimate the statistic distribution based on frequencies. For comparison, we show the classical $G$ statistic. We visualize the distribution in Figure~\ref{fig:distribution-3by3}, Figure~\ref{fig:distribution-5by5} and Figure~\ref{fig:distribution-7by7}.  As we can see from Figure 4, with sufficient data, i.e., $1000$, both $BF_{chi2}$ and $G$ follow $\chi^2$ distribution. Under insufficient data, the probability of $BF_{chi2}$ statistic tends to be overestimated with $\chi^2$ distribution bias towards independence declaration. Compared to uniform prior, $BF_{chi2}$ statistic with Jeffrey's prior produces the distribution that is closer to the $\chi^2$ distribution.

\subsection*{E. Theoretical Guarantee on Bayesian Estimation Improvement over MLE}
To theoretically prove that the Bayesian estimation gives better estimation then MLE, we compare the uncertainty of two estimations via measuring their variances. We consider the parameter estimation. Given data $D$, the closed-form solution for MLE is $\theta^{MLE}_i=\frac{n_i}{N}$, where $n_i$ is the number of samples for state $i$, with $\sum_{i=1}^Kn_i=N$. The variance is then computed as
\begin{equation}
\label{eq:variance-mle-parameters}
    Var(\theta_i^{MLE}) = Var(\frac{n_i}{N}) = \frac{Var(n_i)}{N^2} 
\end{equation}
The variance computed with Eq.~\ref{eq:variance-mle-parameters} captures the variance of estimator caused by the randomness of data. To better show this point, we start from the definition of the variance of the MLE by considering the randomness of data:
\begin{equation}
    Var(\theta^{MLE}) = \int_D (\theta^{MLE} - E[\theta^{MLE}])^2p(D)dD
\end{equation}
Since probability parameters are independent, we can simply compute each of them $\theta^{MLE}_i$ separately, i.e.,
\begin{equation}
\begin{split}
    &Var(\theta^{MLE}_i) \\
    &= \int_D (\theta^{MLE}_i - E[\theta^{MLE}_i])^2p(D)dD\\
    &=\int_D ((\theta^{MLE}_i)^2 -2 \theta^{MLE}_iE[\theta^{MLE}_i] + E^2[\theta^{MLE}_i])p(D)dD
\end{split}
\end{equation}
For MLE estimation, we have $\theta^{MLE}_i = \frac{n_i}{N}$, and $E[\theta^{MLE}_i]=\frac{E[n_i]}{N}$ which is independent of data $D$. The variance of $\theta^{MLE}_i$ then becomes
\begin{equation}
\begin{split}
    Var(\theta^{MLE}_i) &=\int_D (\frac{n_i^2}{N^2} -2\frac{n_i}{N}\frac{E[n_i]}{N}+\frac{E^2[n_i]}{N^2})p(D)dD\\
    &=\int_D \frac{(n_i^2-2n_iE[n_i]+E^2[n_i])}{N^2}p(D)dD\\
    &= \frac{\int_D (n_i-E[n_i])^2p(D)dD}{N^2}\\
    &=\frac{Var(n_i)}{N}
\end{split}
\end{equation}
Given the fact that $n_i$ follows the multinomial distribution with GT parameter $\theta_i$, the variance is $Var(n_i)=N\theta_i(1-\theta_i)$. In the end, we have
\begin{equation}
    Var(\theta_i^{MLE}) = \frac{N\theta_i(1-\theta_i)}{N^2} 
\end{equation}
Similarly, given the closed-form solution for Bayesian estimation is $\theta^{Bayes}_i=\frac{n_i + \alpha_i}{N + \sum_{i}\alpha_i}$, we compute the variance as
\begin{equation}
    \begin{split}
    &Var(\theta_i^{Bayes}) = Var(\frac{n_i + \alpha_i}{N + \sum_{i}\alpha_i}) \\
        &= \frac{Var(n_i + \alpha_i)}{(N + \sum_{i}\alpha_i)^2} = \frac{Var(n_i)}{(N + \sum_{i}\alpha_i)^2}
    \end{split}
\end{equation}
where the variance is $Var(n_i)=N\theta_i(1-\theta_i)$. Furthermore, for hyper-parameters $\alpha_i$, we always have $\alpha_i >0$. In the end, we have
\begin{equation}
\begin{split}
   &Var(\theta_i^{MLE}) \\
   &= \frac{N\theta_i(1-\theta_i)}{N^2} > \frac{N\theta_i(1-\theta_i)}{(N + \sum_{i}\alpha_i)^2} \\
   &= Var(\theta_i^{Bayes})
\end{split}
\end{equation}
Here we show that variance of Bayesian estimation $\theta_i^{Bayes}$ is always smaller than the variance of MLE $\theta_i^{MLE}$ for parameter estimation. 


\subsection*{F. Evaluations on Constraint-based Causal Discovery}
\subsubsection*{F.1. Experiment Settings on Algorithms and Dataset}

\paragraph{Statistical information on datasets:} We employ six benchmark datasets that are widely used for causal discovery evaluation: CHILD, INSURANCE, ALARM, HAILFINDER, CHILD3 and CHILD5. CHILD, INSURANCE and ALARM are medium networks with the number of variables being 20,27 and 37 respectively. HAILFINDER, CHILD3 and CHILD5 are larger and more challenging networks with the number of variables being 56, 60 and 100 respectively. Their information is shown in Table~\ref{tb:BNinfo}. 
\begin{table}[ht!]
\centering
    \adjustbox{max width=0.45\textwidth}{
  \begin{tabular}{|c|c|c|c|c|}
    \hline
    \hline
    Dataset & $\#$Variables &$\#$Edges & Maximum $\#$States \\ 
    \hline
    CHILD & 20 & 25 &6 \\
    INSURANCE & 27 & 52 &5 \\
    ALARM & 37 & 46 & 4 \\
    HAILFINDER
    & 56 & 66 &11 \\
    CHILD3 & 60 &75 &6\\
    CHILD5 & 100 &125 &6\\
    \hline
  \end{tabular}}
  \caption{Benchmark DAG Information}
  \label{tb:BNinfo}
\end{table}

\paragraph{Algorithm setting for local causal discovery:} For the local causal discovery, we employ Causal Markov Blanket (CMB)~\cite{gao2015local}, which is the state-of-the-art method. For hyper-parameters, CMB applies $G$-test and the significance level is set to be $5\%$ as suggested. For $cI^{eB}$, threshold is set based on empirical analysis with synthetic data\footnote{Threshold is estimated as a function of the number of samples per each configuration using synthetic data.}. For $cBF_{chi2}$, we apply the Jeffreys prior ($\alpha^0=\alpha^1=0.5$) and significance level is $5\%$. To verify the performance of original CMB algorithm, we perform CMB on ALARM dataset with 5000 samples. The averaged SHD is 1.63, which is comparable with the reported result, i.e., SHD=1.81~\cite{gao2015local}. 

\paragraph{Algorithm setting for global causal discovery:} We employ RAI as our baseline algorithm and incorporate the proposed independence tests. For hyper-parameters, RAI-BF applies BF with Jeffreys prior for independence test and the threshold is set to 1. PC-stable applies $\chi^2$ independence test, and the significance level is $5\%$ by default. For $rI^{eB}$, threshold is determined based on our empirical analysis with synthetic data. For $rBF_{chi2}$ test, we apply the Jeffreys prior and the significance level is $5\%$. To compare to RAI-BF, we implement BF independence test and incorporate it into RAI\footnote{https://github.com/benzione/FixRAI.}. To verify the implementation, we perform RAI-BF on ALARM dataset with 10,000 samples. The averaged SHD is 25.2, which is comparable to the 25.3 in their reported result~\cite{natori2017consistent}. For PC-stable, we directly apply the algorithm provided by bnlearn\footnote{https://www.bnlearn.com/documentation/man/constraint.html.}. 

\subsubsection*{F.2. Running Time on CPU}
All the experiments in the paper are performed on a laptop with a 2.3 GHz 8-Core Intel Core i9 processor using CPU only. Specifically, we compare the running time of $rI^{eB}$, $rBF_{chi2}$ and RAI-BF since they are all based on the RAI algorithm. We report the average running time over different sample sizes for each dataset. From the results shown in Table~\ref{tab:cputime}, we can see that $rBF_{chi2}$ requires less running time than $rI^{eB}$ and RAI-BF, particularly on large graphs. Though the number of independence tests performed by $rI^{eB}$ is the smallest, the running time per independence test is large, and hence the total running time for $rI^{eB}$ is not competitive.  We will incorporate detailed results into the paper. 
\begin{table}[ht!]
    \centering
    \begin{tabular}{|c|c|c|c|}
    \hline
         Datasets &  $rI^{eB}$ & $rBF_{chi2}$ & RAI-BF\\
         \hline
         CHILD & 5.83s & 1.09s & 0.72s\\
         INSURANCE& 10.08s & 1.53s & 0.80s\\
        ALARM& 24.05s & 2.73s & 1.63s \\
        HAILFINDER& 52.96s & 9.67s& 57.07s \\
        CHILD3& 68.86s & 12.67s & 18.06s \\
        CHILD5& 181.59s & 23.82s& 37.01s \\
        \hline
    \end{tabular}
    \caption{Running Time on CPU}
    \label{tab:cputime}
\end{table}

\subsubsection*{F.3. Performance of RAI-BF with tuned threshold} Since the proposed $BF_{chi2}$ essentially is only an approximate of original $BF$ and thus the BF with the optimal threshold, in principle, should outperform $BF_{chi2}$. But selecting the optimal threshold for the BF can be challenging and incorrect thresholds can lead to inferior performance.  $BF_{chi2}$, in contrast, only needs a default significant level (5$\%$) to perform the test.  This may explain why $BF_{chi2}$ outperforms BF in our experiments. 

Instead of fixing the threshold for RAI-BF with its default value,  we tune its threshold and report the best SHD for comparison. The threshold is selected from [0.2, 2.0]. The corresponding number of independence tests is also reported. From the results shown in Table~\ref{tb:optimalRAIBF}, we can see that $rI^{eB}$ achieves overall better efficiency by performing a smaller number of independence tests, which is consistent with our conclusion stated in the paper. Comparing the RAI-BF with the optimally tuned threshold and $rBF_{chi2}$, they achieve comparable performance in terms of both accuracy and efficiency.  
\begin{table}[ht!]
  \centering
  \adjustbox{max width=0.45\textwidth}{
  \begin{tabular}{|c|ccc|ccc|}
    \hline
    &\multicolumn{3}{c|}{\underline{SHD}}&\multicolumn{3}{c|}{\underline{\#Independence Test}}\\
    Dataset &$rI^{eB}$ & $rBF_{chi2}$ & RAI-BF&$rI^{eB}$ & $rBF_{chi2}$ & RAI-BF\\ 
    \hline
    CHILD& 19.7 & 19.3  & 19.4& 350 &538  &493\\
    INSURANCE& 48.6 & 44.7 & 45.1& 575 &930 &757\\
ALARM& 41.7 & 36.2 & 32.9&1161 & 1576 &1708\\
HAILFINDER& 88.4 & 104.3 & 103.9&2215 &3129 &2531\\
CHILD3& 65.0 & 53.2 & 63.4&3786 & 4518 & 2260\\
CHILD5& 120.0 & 106.8 & 116.7&9197 & 9857 &8028\\
    \hline
  \end{tabular}}
  \caption{Comparison to optimal RAI-BF with tuned threshold}
  \label{tb:optimalRAIBF}
\end{table}
On Child3 and Child5, RAI-BF, with an optimally selected threshold, performs a smaller number of independence tests as the optimal threshold is small ($\sim$0.2), leading to more independence declarations.  

\subsubsection*{F.4. Performance Improvement Consistency with PC and MMHC}
To further show that our proposed methods can consistently improve the causal discovery performance, we consider another two widely used DAG learning algorithms: PC~\cite{spirtes2000causation} and MMHC~\cite{tsamardinos2006max}. We incorporate the proposed methods into PC and MMHC for evaluation. 
PC with empirical Bayesian MI estimation and $BF_{chi2}$ independence test are denoted as $pcI^{eB}$ and $pcBF_{chi2}$ respectively. MMHC with empirical Bayesian MI estimation and $BF_{chi2}$ independence test are denoted as $mI^{eB}$ and $mBF_{chi2}$ respectively. 
\begin{table}[ht!]
  \centering
    \adjustbox{max width=0.45\textwidth}{
  \begin{tabular}{|c|ccc|ccc|}
    \hline
    \hline
    Method&\multicolumn{6}{c||}{\textbf{PC}}\\ \hline
    Dataset&\multicolumn{3}{c|}{\underline{SHD}}&\multicolumn{3}{c||}{\underline{\#Independence Test}}\\ 
    (\textbf{MEAN}) &$pcI^{eB}$ & $pcBF_{chi2}$ & PC&$pcI^{eB}$ & $pcBF_{chi2}$ & PC \\ \hline
    CHILD &\textbf{22.0} & \textbf{22.0} & 27.4 & \textbf{331} & 382 & 610\\
    INSURANCE & 50.4&\textbf{50.0}&57.4 & \textbf{548} & 699 & 1067\\
    ALARM & 41.9 & \textbf{40.5} & 58.2 & \textbf{1098} & 1355 & 3585\\
    HAILFINDER &\textbf{84.1} &101.1  &119.0&\textbf{2511}&3959  &33352\\
    CHILD3 & 77.2 & \textbf{75.2} & 88.8 & \textbf{2099} & 2548 & 3880\\ 
    CHILD5 & 108 &\textbf{95.3}  &107.8 &13180  &11009  &\textbf{10348}\\\hline
    \end{tabular}}
    	\caption{Evaluation of proposed methods with PC}
  \label{tb:pc}
\end{table}
\begin{table}[ht!]
  \centering
    \adjustbox{max width=0.45\textwidth}{
  \begin{tabular}{|c||ccc|ccc|}
    \hline
    \hline
    Method&\multicolumn{6}{c|}{\textbf{MMHC}}\\ \hline
    Dataset&\multicolumn{3}{c|}{\underline{SHD}}&\multicolumn{3}{c|}{\underline{\#Independence Test}}\\ 
    (\textbf{MEAN}) &$mI^{eB}$ & $mBF_{chi2}$ & MMHC&$mI^{eB}$ & $mBF_{chi2}$ & MMHC\\ \hline
    CHILD &22.5 & \textbf{21.9} & 22.4 & \textbf{14} & \textbf{14} & 37\\
    INSURANCE &52.1&\textbf{49.9}&52.9 & \textbf{19} & 23 & 45\\
    ALARM & 42.0 & \textbf{38.7} & 40.1 & \textbf{24} & 26 & 33\\
    HAILFINDER &\textbf{78.3}  &87.0  &92.0 &10  &\textbf{5}  &34\\
    CHILD3 & 66.9 & \textbf{63.7} & 64.9 & \textbf{8} & 20 & 24\\ 
    CHILD5 &103  &\textbf{102}  &104 &\textbf{9}  &25  &24\\\hline
    \end{tabular}}
    	\caption{Evaluation of proposed methods with MMHC}
  \label{tb:mmhc}
\end{table} 

As shown in Table~\ref{tb:pc} and Table~\ref{tb:mmhc}, our proposed methods can consistently improve the DAG learning performance, particularly with PC. For example, on ALARM, $pcBF_{chi2}$ achieves averaged SHD $40.5$, while PC only achieves averaged SHD $58.2$. MMHC is a hybrid approach, where a constraint-based algorithm is only used to obtain an initial graph for a score-based algorithm. Thus, the performance of MMHC doesn't completely reflect the performance of independence tests, and 
different independence tests don't introduce much difference to DAG learning performance. Overall, $BF_{chi2}$ achieves better accuracy and $I^{eB}$ achieves better efficiency with both PC and MMHC on different datasets. 

\subsection*{G. Evaluation of Independence Tests}

\subsubsection*{G.1. Evaluation on Synthetic Data}
We perform experiments on synthetic data to study the performance of the proposed independence tests. We firstly evaluate
the proposed empirical Bayesian MI estimation. We then analyze the proposed $BF_{chi2}$ independence test. For the synthetic data, we consider two multi-state random variables $X$ and $Y$. The underlying probabilistic dependency between $X$ and $Y$ is randomly generated. The probability parameters are randomly generated given the dependency with the symmetric Dirichlet prior. We generate synthetic data of different small sizes for evaluation.

\paragraph{Mutual Information Estimation} We compare our proposed empirical Bayesian MI estimation with state-of-the-art MI estimation methods. Specifically, we consider the adaptive partition method~\cite{article},
the empirical Bayesian method~\cite{hutter2002distribution} and the full Bayesian method~\cite{archer2013bayesian}. 
To measure the accuracy of the MI estimation, 
we apply the absolute error between the estimated MI $\hat{MI}$ and the ground truth MI $MI$, i.e., $|\hat{MI}-MI|$. We report the averaged absolute error over 1000 runs for each sample size. 
\begin{figure}[!ht]
   \centering
   \includegraphics[width= 0.45\textwidth]{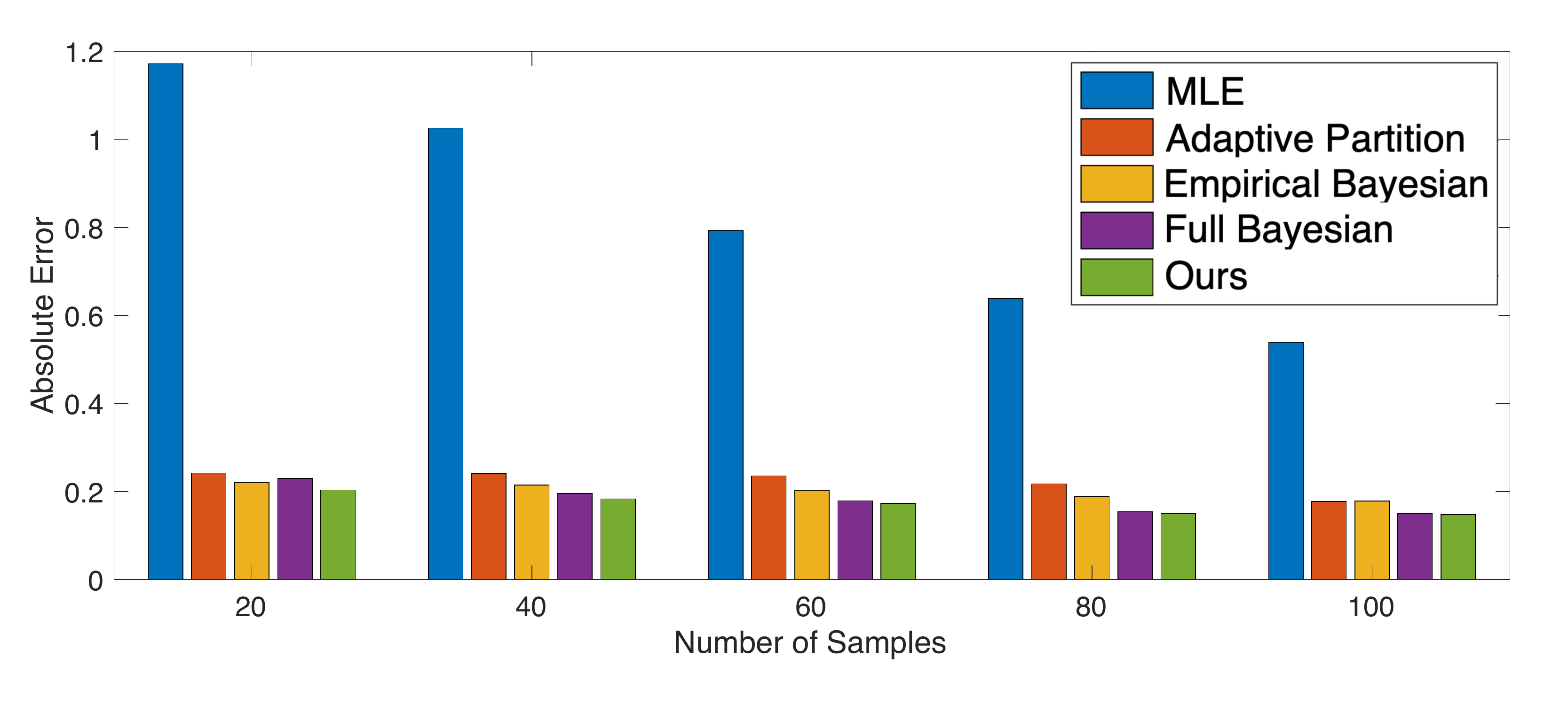}
   \caption{Mutual Information Estimation}
   \label{fig-MIestimation}
\end{figure}
From Figure~\ref{fig-MIestimation}, we can see that our approach gives the best estimation compared to others.  In particular, our proposed empirical Bayesian MI via a MAP estimation of the hyper-parameter $\alpha$ performs better than the empirical Bayesian method with fixed $\alpha$~\cite{hutter2002distribution}. Additionally, we achieve comparable accuracy compared to the full Bayesian method~\cite{archer2013bayesian}. Without requiring the integration over the hyper-parameter space, our methods only takes $0.002$ seconds on average to finish one run, while the full Bayesian method~\cite{archer2013bayesian} needs on average $10.481$ seconds to finish one run. Thus, our empirical Bayesian MI estimation is more computationally efficient to be applied in causal discovery.

\paragraph{Hypothesis testing based Independence Test} We compare the proposed $BF_{chi2}$ independence test against the standard $G$ test and the Bayes Factor (BF)~\cite{natori2017consistent} which represents the state-of-the-art independence test that matches with our approach.
We follow the BF~\cite{natori2017consistent} and apply Jeffreys prior. We consider Type-1 error and Power as measurements. Type-1 error is rejecting the null hypothesis $H_0$ when it is true.
\begin{figure}[ht]
    \centering
    \includegraphics[width=.45\textwidth,height=1.4in]{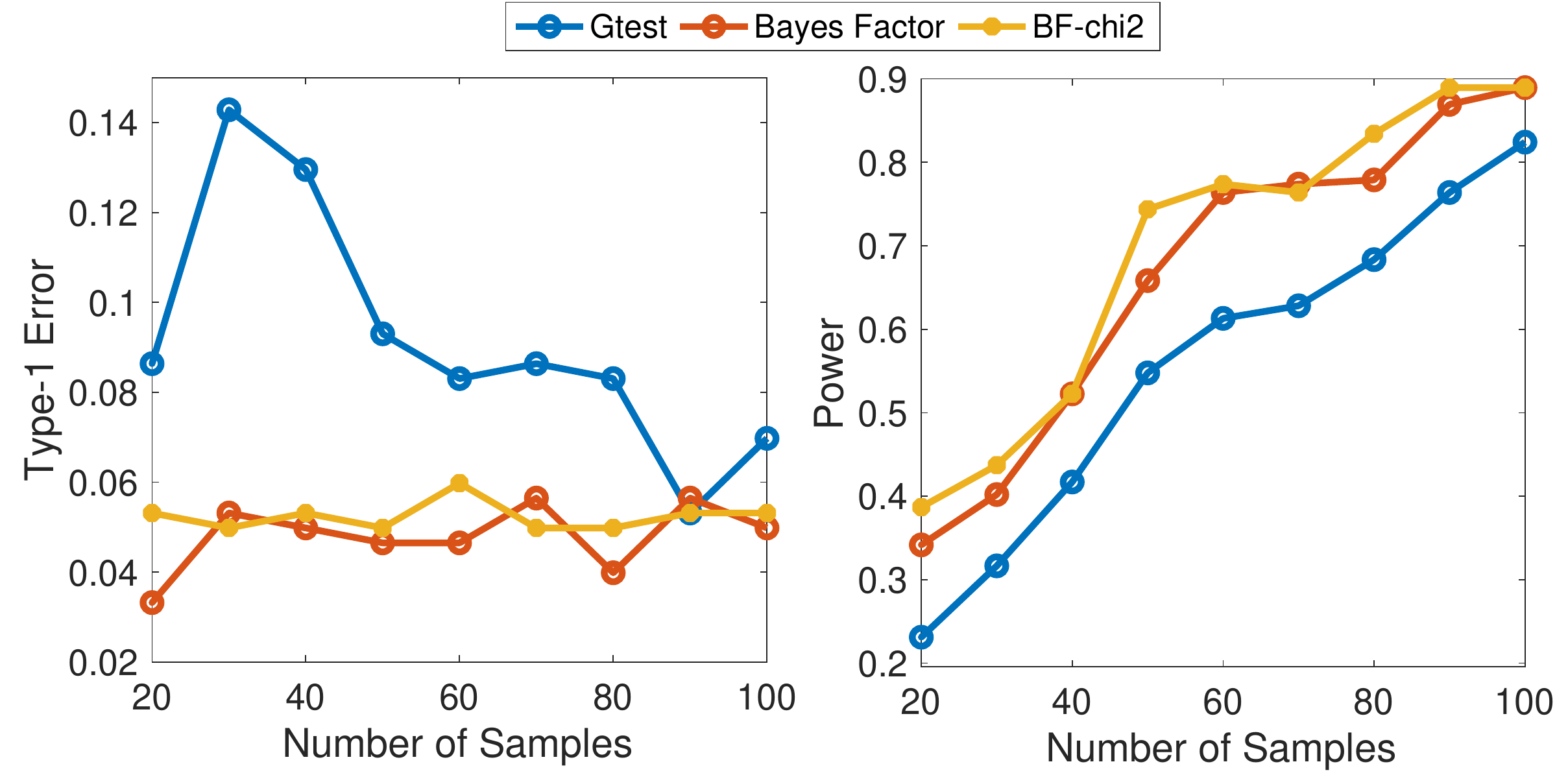}
    \caption{Performance of BF-chi2}
    \label{fig:BF-chi2}
    \vspace*{-0.6\baselineskip}
\end{figure}
The power is the probability of correctly rejecting $H_0$. We set the significance level of $BF_{chi2}$ test, $G$ test, and
manually tune the threshold of BF to make their corresponding Type-1 error near $5\%$. And we compare the power of different methods. 
From the results shown in Figure~\ref{fig:BF-chi2}, we can see that Bayesian approaches $BF_{chi2}$ and BF achieve higher power than frequentist-based approach $G$ test under insufficient data with lower Type-1 error. And our approach achieves slightly better power compared to the BF~\cite{natori2017consistent}, without any threshold tuning.

\subsubsection*{G.2. Evaluation on Benchmark Datasets} We compare proposed independence tests to two state-of-the-art methods: adaptive partition and
empirical Bayesian with fixed $\alpha$ methods in terms of causal discovery performance on benchmark datasets. We incorporate the adaptive partition method and the empirical Bayesian with fixed $\alpha$ method to RAI (denoted as $rI^{AdP}$ and $rI^{eBFix}$ respectively). 
\begin{table}[ht!]
  \centering
  \scalebox{0.8}{
  \begin{tabular}{|c|ccccc|}
    \hline
    \hline
    Dataset&\multicolumn{5}{c|}{\underline{\#Independence Test}}\\ 
    (\textbf{MEAN}) &$rI^{AdP}$ & $rI^{eBFix}$ & $rI^{eB}$&$rBF_{chi2}$ & RAI-BF \\ \hline
    CHILD & \textbf{267} & 552 & 350 & 538 & 955\\
    INSURANCE & \textbf{516} & 589 & 575 & 930 & 1012\\
    ALARM & \textbf{915} & 1236 & 1161 & 1576 & 1857\\
    HAILFINDER &\textbf{1806}&2486&2215&3129&10212\\
    CHILD3 & \textbf{2642} & 4299 & 3786 & 4518 & 5503\\ 
    CHILD5 &\textbf{7278}&9889&9197&9857&11174\\\hline
    \end{tabular}}
    	\caption{Efficiency comparison to SoTA independence tests}
  \label{tb:CItestsEfficiency}
\end{table}
In terms of efficiency, as shown in Table~\ref{tb:CItestsEfficiency}, $rI^{AdP}$ achieves best efficiency by performing the smallest number of independence tests. Though $rI^{AdP}$ is efficient, its accuracy is compromised and $rI^{AdP}$ performs worse than other methods on almost all the datasets. Our proposed method $rI^{eB}$ is the second most efficient method, and achieves competitive accuracy at the same time.
Overall, our proposed methods outperform other SoTA independence tests in terms of causal discovery performance. On HAILFINDER, because $rI^{AdP}$ tends to declare independence, the learned DAG contains fewer false positive edges compared to other methods and thus its averaged SHD is the best. 

\subsection*{H. Classification Performance under Imbalanced Data} 
To further demonstrate the effectiveness of the proposed methods, we consider real world imbalanced datasets where samples for certain classes are insufficient. Particularly, we construct a structured classifier by using the learned DAG from global causal discovery methods. We perform the evaluation on four UCI datasets~\cite{Dua:2019} that are benchmark imbalanced datasets~\cite{jiang2014cost} \cite{Jiang2013Sampled}. The statistic information of datasets is shown in Table~\ref{tb:UCIinfo}. 
\begin{table}[ht!]
\centering
	\caption{Information of UCI datasets}
  \label{tb:UCIinfo}
  \adjustbox{max width=0.45\textwidth}{
  \begin{tabular}{|c|c|c|c|c|}
    \hline
    \hline
    Dataset & $\#$Samples &$\#$Attributes &$\#$Majority/$\#$Minority \\ \hline
    Breast-w & 699 & 9 & 458/241\\
    Spect & 267 & 22 & 212/55\\
    Diabetes & 768 & 8 & 500/268\\
    Parkinsons & 195 & 23 & 147/48\\
    \hline
  \end{tabular}}
\end{table}
F1-score and AUC are applied as the evaluation metrics to measure the classification accuracy. We apply 3-fold cross-validation. The settings of hyper-parameters for independence tests remain the same as we did in the previous section. Results are shown in Table~\ref{tb:accglobal_imbalance}.

\begin{table}[ht!]
\centering
	\caption{Predictive performance under imbalanced data}
  \label{tb:accglobal_imbalance}
    \adjustbox{max width=0.45\textwidth}{
  \begin{tabular}{|c|cccc|}
    \hline
    \hline
    &\multicolumn{4}{c|}{\underline{F1-score / AUC}}\\
    Dataset &$rI^{eB}$ & $rBF_{chi2}$ & RAI-BF & PC-stable\\ 
    \hline
    Breast-w &0.42/\textbf{0.52} & \textbf{0.84}/0.50 & 0.40/0.49 & 0.47/0.24 \\ 
    Spect &\textbf{0.59}/0.74 &0.58/\textbf{0.77} & 0.57/\textbf{0.74} & 0.41/\textbf{0.55} \\ 
    Diabetes & 0.63/0.44 &\textbf{0.68}/\textbf{0.43} &0.50/\textbf{0.51} & 0.67/\textbf{0.44} \\
    Parkinsons &0.61/\textbf{0.74} &\textbf{0.70}/0.64 & 0.38/0.50 & 0.48/0.54\\ 
    \hline
  \end{tabular}}
\end{table}

We can see that $rBF_{chi_2}$ achieves better performance on most of the datasets, and $rI^{eB}$ also achieves improved performance. For example, on Breast-w dataset, $rBF_{chi_2}$ improves the F1-score by $44\%$ and $37\%$  compared to RAI-BF method and PC-stable method respectively. Considering AUC, we can see that both $rI^{eB}$ and $rBF_{chi2}$ achieve better AUC than baseline methods RAI-BF and PC-stable on most of the datasets. These results further demonstrate that, with proposed independence tests, we can learn DAGs that better capture underlying structure among variables under imbalanced data, leading to improved structure classification performance.

\bibliographystyle{named}
\bibliography{egbib}


\end{document}